\documentclass{article}

\usepackage{arxiv}

\usepackage[utf8]{inputenc} 
\usepackage[T1]{fontenc}    
\usepackage{hyperref}       
\usepackage{url}            
\usepackage{booktabs}       
\usepackage{amsfonts}       
\usepackage{nicefrac}       
\usepackage{microtype}      
\usepackage{lipsum}         
\usepackage{graphicx}
\usepackage{doi}
\usepackage{amsmath, amssymb}
\usepackage{bbm}
\usepackage{array}          
\usepackage{multirow}
\usepackage[dvipsnames]{xcolor}
\usepackage[numbers]{natbib}

\DeclareUnicodeCharacter{202F}{\,}

\title{Edge-Based Multimodal Sensor Data Fusion with Vision Language Models (VLMs) for Real-time Autonomous Vehicle Accident Avoidance}

\author{ \href{https://orcid.org/0009-0000-8734-5498}{\includegraphics[scale=0.06]{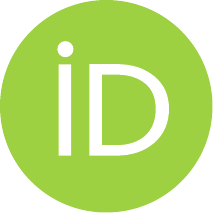}\hspace{1mm}Fengze~Yang} \\
	Department of Civil \& Environmental Engineering\\
	University of Utah\\
	201 Presidents' Cir, \\
    Salt Lake City, UT 84112, USA \\
	\texttt{fred.yang@utah.edu} \\
    \And
	\href{https://orcid.org/0009-0003-0211-9953}{\includegraphics[scale=0.06]{orcid.pdf}\hspace{1mm}Bo Yu}\\
	Department of Civil \& Environmental Engineering\\
	University of Utah\\
	201 Presidents' Cir, \\
    Salt Lake City, UT 84112, USA \\
	\texttt{bo.yu@utah.edu} \\
	\And
	\href{https://orcid.org/0000-0001-5366-5389}{\includegraphics[scale=0.06]{orcid.pdf}\hspace{1mm}Yang Zhou}\\
	Zachry Department of Civil and Environmental Engineering\\
	Texas A\&M University\\
	199 Spence Street, Room 301A, \\
    College Station, TX 77840, USA \\
	\texttt{yangzhou295@tamu.edu} \\
	\And
	\href{https://orcid.org/0009-0006-7737-1178}{\includegraphics[scale=0.06]{orcid.pdf}\hspace{1mm}Xuewen Luo}\\
	Department of Civil \& Environmental Engineering\\
	University of Utah\\
	201 Presidents' Cir, \\
    Salt Lake City, UT 84112, USA \\
	\texttt{u1594388@utah.edu} \\
	\And
	\href{https://orcid.org/0000-0002-7594-2292}{\includegraphics[scale=0.06]{orcid.pdf}\hspace{1mm}Zhengzhong Tu}\\
	Department of Computer Science \& Engineering\\
	Texas A\&M University\\
	400 Bizzell St, \\
    College Station, TX 77840, USA \\
	\texttt{zhengzhong.tu@utexas.edu} \\
	\And
	\href{https://orcid.org/0000-0002-5447-4768}{\includegraphics[scale=0.06]{orcid.pdf}\hspace{1mm}Chenxi~Liu}\thanks{Corresponding author.} \\
	Department of Civil \& Environmental Engineering\\
	University of Utah\\
	201 Presidents' Cir, \\
    Salt Lake City, UT 84112, USA \\
	\texttt{chenxi.liu@utah.edu} \\
}

\date{}

\renewcommand{\shorttitle}{\textit{arXiv} Template}

\hypersetup{
pdftitle={A template for the arxiv style},
pdfsubject={q-bio.NC, q-bio.QM},
pdfauthor={David S.~Hippocampus, Elias D.~Striatum},
pdfkeywords={Traffic safety, Edge AI, Autonomous Driving, Vision-Language Model, V2X Communication},
}

\begin{document}
\maketitle

\begin{abstract}
Autonomous driving (AD) systems relying solely on onboard sensors may fail to detect distant or obstacle hazards, potentially causing preventable collisions; however, existing transformer-based Vehicle-to-Everything (V2X) approaches, which mitigate AD sensing limitations, either lack effective multimodal fusion and reasoning or struggle to meet real-time performance requirements under complex, high-dimensional traffic conditions. This paper proposes the \textbf{R}eal-time \textbf{E}dge-based \textbf{A}utonomous \textbf{C}o-pilot \textbf{T}rajectory planner (REACT), a V2X-integrated trajectory optimization framework for AD based on a fine-tuned lightweight Vision-Language Model (VLM). REACT integrates infrastructure-provided hazard alerts with onboard sensor data, capturing intricate surrounding traffic dynamics and vehicle intents through visual embeddings, interpreting precise numerical data from symbolic inputs, and employing contextual reasoning to generate optimized, safety-oriented trajectories. To ensure robust real-time deployment on edge devices, REACT innovatively employs Residual Trajectory Fusion (RTF) design and specialized edge-adaptation strategies to reduce model complexity and improve inference efficiency. Evaluated on the DeepAccident benchmark, REACT achieves state-of-the-art performance, a 77\% collision rate reduction, a 48.2\% Video Panoptic Quality (VPQ), and a 0.57-second inference latency on the Jetson AGX Orin. Ablation studies validate the contribution of each input, module, and edge adaptation strategy. These results highlight the effectiveness of lightweight VLMs in enabling real-time cooperative planning on edge platforms and underscore the potential of language-guided contextual reasoning for improving traffic safety and responsiveness.

\end{abstract}


\section{Introduction}

Road traffic crashes remain a critical global issue, claiming approximately 1.19 million lives and injuring up to 50 million individuals annually. They are the leading cause of death among individuals aged 5 to 29 \cite{who_road_traffic_injuries}. Among the contributing factors, impaired, distracted, or careless driving stands out as the primary cause, accounting for 29.2\% of all accidents and resulting in over 3,000 fatalities each year in the U.S. alone \cite{nhtsa2023, cdc_distracted_driving}. This significant toll underscores the urgent need for autonomous driving (AD) and Advanced Driver-Assistance Systems (ADAS) to help mitigate human error and reduce distraction-related collisions \cite{zhang2025anticipatory}. Although these technologies hold promise for enhancing road safety, their capabilities are often limited by the perception range of onboard sensors, which can delay or prevent the detection of occluded or distant hazards. As a result, ADAS and AD systems may fail to respond effectively to unexpected threats, highlighting the necessity for a more proactive and context-aware driving framework \cite{huang2023v2x, li2025simulating}.

To overcome the safety limitations of isolated onboard sensing, connected autonomous vehicle (CAV) networks leverage Vehicle-to-Everything (V2X) communication to integrate diverse transportation agents, including vehicles, roadside units (RSUs), and traffic management entities. This collaborative framework enhances system-wide situational awareness, enabling individual vehicles to perceive beyond their immediate sensor range, including blind spots and broader traffic conditions, thereby supporting more proactive and informed decision-making \cite{abdi2024advancing}. Despite its potential, large-scale V2X deployment faces notable challenges in prevailing perception methodologies. Models such as Convolutional Neural Networks (CNNs), Recurrent Neural Networks (RNNs), and Vision Transformers (ViTs) \cite{khan2022transformers} often suffer from limited generalization, requiring extensive labeled datasets and struggling with novel or unseen scenarios \cite{raghu2021vision, cui2025cmoa}. Moreover, non-attention-based methods excel primarily in pattern recognition but fail to meet the high-level contextual reasoning required for complex real-world driving \cite{wang2023decision, fu2024summary}. Besides the aforementioned limits, integrating multimodal information remains a substantial challenge due to the architectural complexity and limitations of existing model designs \cite{joshi2021review, truhn2024large}.

Recent advances in Vision-Language Models (VLMs), extensions of Large Language Models (LLMs) with integrated visual processing capabilities \cite{brown2020language}, present promising solutions to overcome the limitations of conventional perception systems \cite{wang2025generative, pu2024frontiers}. Modern traffic environments are dynamic, complex, and context-rich \cite{zhang2025virtual}, demanding autonomous systems that can understand and respond to diverse multimodal inputs. VLMs address this challenge by combining semantic reasoning with visual context, enabling more informed and adaptive decision-making, outperforming conventional models \cite{wang2025generative}. First, VLMs excel in high-level contextual reasoning, interpreting spatiotemporal cues, anticipating agent intent, and understanding implicit traffic rules, which are essential for navigating scenarios like occluded merges, crowded intersections, or rare accident scenes \cite{wang2024exploring, yang2025self}. Second, VLMs inherently support multimodal integration, effectively fusing heterogeneous sensor data and textual inputs \cite{li2024vision}, which facilitates comprehensive perception and enables more accurate, adaptive, and timely decision-making in real-world driving scenarios \cite{you2024v2x, luo2025v2x}. Third, VLMs demonstrate strong generalization through zero-shot learning \cite{saha2024improved, lei2024ez}, reducing the dependency on large-scale labeled datasets and enhancing the robustness and reliability of perception systems. Finally, their ability to pair linguistic explanations with visual perception improves model interpretability \cite{hassija2024interpreting, xu2019explainable}, supporting more efficient human-vehicle interactions by winning the trust of human drivers.  

Building upon these insights, this research introduces a Real-time Edge-based Autonomous Co-pilot Trajectory planner (REACT), a V2X-enabled framework leveraging a lightweight VLM to improve transportation safety. REACT comprises five core modules: 1) sensing data acquisition, 2) data structuring, 3) task alignment, 4) task projection enhancement, and 5) Residual Trajectory Fusion (RTF). The sensing module gathers multimodal inputs, including onboard camera frames, navigation goals, ego vehicle states, and hazard alerts from Roadside Units (RSUs). Visual data is converted into Bird’s-Eye View (BEV) maps that capture road geometry and nearby road users. To support VLM's motion reasoning, REACT stacks BEV maps from the current moment and a prior timestamp, allowing the VLM to infer agent trajectories and intentions. The data structuring module filters noise, extracts necessary information, and validates the RSU hazards. The task alignment stage aligns input coordination and timestamps, then encodes both visual and symbolic information into optimized prompts. The task projection enhancement module then passes those prompts into a lightweight VLM, which, guided by Chain-of-Thought prompting, predicts necessary trajectory corrections. Finally, the RTF module applies these residuals to the nominal trajectory, yielding a refined path that proactively avoids potential collisions.

A notable feature of REACT is its edge-based deployment, which ensures real-time responsiveness and minimizes delays caused by network latency and bandwidth limitations with edge computing \cite{chen2023edge, baller2021deepedgebench}. VLMs demand substantial computational resources, often exceeding the capabilities of edge devices \cite{chen2023edge}. To address this challenge, REACT introduces innovative optimization strategies that combine prompt engineering \cite{marvin2023prompt} with fine-tuning techniques on the lightweight model \cite{chen2024overview, luo2024delving}, achieving a model reaction latency reduction of 93.67\% without sacrificing performance. In terms of model architecture, REACT incorporates a novel module, RTF, which leverages original trajectory data to enhance the outputs of the VLM. The RTF module is compact, modular, and readily adaptable for deployment in other trajectory prediction tasks. Experimental results indicate that the combination of REACT’s optimization strategies and the RTF module improves system efficiency by 95.25\%, enhances trajectory prediction accuracy, and ensures greater training stability.

In summary, REACT addresses key limitations of conventional perception methodologies in V2X systems by deploying a lightweight VLM on edge devices. The primary contributions of this research are as follows:

\begin{itemize}
    \item \textbf{Architectural Contribution:} This paper introduces REACT, a V2X-integrated autonomous trajectory planning framework built upon a customized VLM to enable safer and more context-aware autonomous driving.

    \item \textbf{Methodological Innovation:} This study designs a lightweight VLM-based sensor fusion system that integrates multimodal information via spatial-temporal alignment and spatial grounding enhancement, with only 512M parameters. Furthermore, an innovative structure Residual Trajectory Fusion (RTF) is proposed to improve both system computation efficiency and accuracy for collision risks mitigation.

    \item \textbf{Edge Optimization:} REACT incorporates targeted edge adaptation strategies that streamline model complexity, enabling real-time deployment on resource-constrained edge devices without compromising predictive quality.

    \item \textbf{Practical Applicability:} The proposed system is validated under diverse traffic scenarios, environmental conditions, and infrastructure configurations, demonstrating robustness and adaptability in real-world deployment settings.
\end{itemize}


\section{Literature Review}
LLMs, or specifically VLMs, are increasingly recognized for their potential to enhance perception and decision-making in AD. To contextualize this research, this review highlights three critical dimensions: the application of VLMs in AD systems, their integration with V2X communication to enable cooperative perception, and their adaptation for edge deployment to support real-time, resource-efficient operations.

\subsection{VLMs in AD Systems}

Recent studies have positioned VLMs as promising tools for enhancing scene understanding, reasoning, and decision-making in AD. Research to date has explored the application of VLMs across a range of AD tasks, including perception, trajectory prediction, planning, and high-level reasoning.

\textit{Scene Perception and Understanding}

VLMs have shown potential in improving situational awareness and semantic interpretation of complex driving environments. DriveVLM \cite{tian2024drivevlm} introduced a VLM-based framework that integrates scene description, analysis, and hierarchical planning modules, demonstrating enhanced performance in long-tail and complex urban scenarios. Similarly, DriveLM \cite{sima2024drivelm} proposed a Graph VQA-based architecture that supports multi-step reasoning to refine perception outputs, effectively mimicking human-like cognitive processes in perception pipelines. AutoTrust \cite{xing2024autotrust} assessed the reliability of VLMs in scene understanding tasks, highlighting persistent vulnerabilities to adversarial conditions and fairness concerns, even in purpose-built models like DriveVLM. Safety-critical event (SCE) understanding is another emerging area where ScVLM \cite{shi2025scvlm} has made notable progress by enhancing VLMs with contrastive learning and classification capabilities, enabling improved detection and interpretation of rare or hazardous scenarios. Collectively, these studies underscore the growing importance of robust perception systems and the need for stronger safety guarantees in VLM-based approaches.

\textit{Trajectory Planning and Prediction}

Beyond perception, VLMs are increasingly applied to trajectory generation and motion planning, enabling semantic intent inference beyond low-level sensor data. HE-Drive \cite{wang2024he} employed VLMs as trajectory scorers within a diffusion-based planner, ensuring smoother, more comfortable, and human-like driving trajectories. Other frameworks, such as VLM-AD \cite{xu2024vlm} and VLM-E2E \cite{liu2025vlm}, incorporated VLMs as auxiliary supervision tools to infuse common-sense reasoning and semantic attention into end-to-end learning models. These methods demonstrated improved trajectory accuracy and planning robustness, while strategically decoupling VLM inference from real-time execution to minimize runtime overhead.

\textit{High-Level Reasoning and Decision-Making}

VLMs are also gaining traction in high-level decision-making tasks, bridging perception and cognition. Both DriveVLM and DriveLM incorporate reasoning layers that interpret perception outputs into planning decisions via natural language. Further, VLM-RL \cite{huang2024vlm} introduced a novel reward-generation mechanism based on language-goal contrast, aligning reinforcement learning policies with semantically interpretable objectives. These approaches illustrate how VLMs can serve as a semantic bridge between low-level observations and high-level decision frameworks, enabling more transparent and adaptive driving behaviors.

\textit{Generalist VLM Models}

Recent efforts have focused on developing unified, multitask models that address diverse AD tasks within a single framework. DriveMM \cite{huang2024drivemm} proposed a large multimodal model trained across perception, prediction, and planning tasks, highlighting the scalability and generalization potential of VLM-centric systems. Likewise, EMMA \cite{hwang2024emma} and SimpleLLM4AD \cite{zhang2024edgeshard} showcased end-to-end VLM-driven pipelines that integrate object detection, scene reasoning, and motion planning within a shared language-based representation space.

VLMs are reshaping the landscape of AD by supporting integrated perception, reasoning, and planning workflows. However, a critical limitation remains that onboard perception is inherently constrained by the sensing range of vehicle-mounted sensors, limiting situational awareness in complex traffic environments. This highlights the need for cooperative perception mechanisms, where VLMs can be further enhanced through integration with V2X communication systems.

\subsection{VLM-V2X Integration}
The integration of VLMs with V2X communication is an emerging research direction with the potential to enhance cooperative perception. Although still in its early stage, recent studies have begun exploring the synergy between semantic language representations and V2X-enabled driving environments. One of the earliest contributions in this domain, V2X-VLM \cite{you2024v2x}, presents an end-to-end V2X cooperative AD framework augmented by VLMs. By fusing multimodal data from both vehicle and infrastructure sensors into a unified semantic representation, the framework enhances trajectory planning and situational awareness under real-world conditions. The incorporation of contrastive learning further improves feature discrimination and system robustness in complex driving scenarios, highlighting the promise of VLM-based semantic fusion for cooperative perception. 

Further research has been demonstrated in frameworks such as the Advanced Driving Agent with Multimodal LLMs (MLLMs) \cite{chen2024advanced} and AutoReward \cite{han2024autoreward}, which emphasize semantic reasoning and high-level abstraction. These systems utilize multimodal inputs and chain-of-thought (CoT) prompting mechanisms, enabling distributed decision-making and adaptive policy learning through semantic cues. SenseRAG \cite{luo2025senserag} builds upon this direction by applying language-based reasoning in a V2X framework. Using retrieval-augmented generation (RAG), it constructs a proactive environmental knowledge base capable of interpreting real-time sensor and V2X data through CoT prompting. This architecture offers a flexible and latency-aware approach to multimodal data fusion.

At a broader conceptual level, LLMs are also laying the groundwork for future VLM-V2X integration. AgentsCoMerge \cite{hu2024agentscomerge} introduces a LLM-driven multi-agent collaborative decision-making system for ramp merging, where semantic communication modules facilitate coordination among agents. Similarly, V2X-LLM \cite{wu2025v2x} integrates LLMs into connected vehicle corridors to support real-time understanding and reasoning over large-scale V2X data streams, including Basic Safety Messages (BSMs) and Signal Phase and Timing (SPaT) information. While these approaches employ general LLMs rather than VLMs, the architectures are well-positioned to incorporate VLM-generated messages in future cooperative driving scenarios.

Despite recent research demonstrating the potential of VLMs in enhancing AD systems, and above explorations showing promise in integrating VLMs with V2X communication for cooperative decision-making, most existing efforts remain limited in scope. Current architectures are often lacking lightweight and real-time implementations suited for deployment in dynamic driving environments. These limitations highlight the pressing need for edge-adapted VLM frameworks that can support scalable, responsive, and resource-efficient integration in practical AD scenarios with V2X communication.

\subsection{VLM Deployment on Edge Devices}
Edge computing enables computational intelligence to be deployed near data sources—such as vehicles or roadside units—supporting low-latency, privacy-preserving, and bandwidth-efficient processing in AD systems \cite{chen2019deep}. As VLMs grow increasingly complex, adapting them for real-time inference on resource-constrained edge devices has become essential for practical deployment in AD environments.

Recent research has proposed various architectural strategies to enhance edge-level deployment of VLMs. EdgeVLA \cite{budzianowski20edgevla} introduces structural simplifications, such as removing autoregressive decoding and utilizing smaller language models, to accelerate visuomotor inference, offering design insights suitable for time-sensitive driving tasks. EdgeCloudAI \cite{ghasemi2024edgecloudai} presents a hybrid architecture where lightweight preprocessing occurs on edge nodes, while deeper semantic reasoning is offloaded to the cloud, achieving scalable and low-latency scene interpretation. Similarly, VEANET \cite{zhong2025generative} proposes a diffusion-based contract mechanism for dynamically migrating AI agent twins across roadside units, enabling continuous edge-assisted support for moving autonomous vehicles.

Complementing these architectural innovations, a growing body of work emphasizes the importance of model compression and performance evaluation under resource constraints. MobileAIBench \cite{murthy2024mobileaibench} provides a benchmarking toolkit to assess the impact of quantization and runtime efficiency on LLMs and MLLMs in mobile environments, revealing trade-offs among latency, computation cost, and task performance. MiniCPM-V \cite{yao2024minicpm} demonstrates that GPT-4V-comparable performance can be achieved on mobile hardware through architectural streamlining and alignment techniques. Likewise, MobileVLM and MobileVLM V2 \cite{chu2023mobilevlm, chu2024mobilevlm} showcase VLMs with 1.7B or 3B parameters delivering competitive results at faster inference speeds on edge platforms. Further, Align-KD \cite{feng2024align} proposes a cross-modal knowledge distillation method that effectively preserves visual-textual alignment accuracy in compressed student models.

Emerging lightweight frameworks reflect a broader shift toward practical, deployable VLM systems. AppVLM \cite{papoudakis2025appvlm} introduces a compact model capable of executing mobile interaction tasks with near-GPT-4o accuracy at a lower computational cost, highlighting trade-offs applicable to in-vehicle assistant systems. MobileExperts \cite{zhang2024mobileexperts} and Mobile-Env \cite{zhang2023mobile} extend this vision by offering modular agent frameworks and standardized evaluation tools to facilitate edge-optimized AV interface development.

While progress has been made in optimizing VLMs for edge deployment, much of the existing work remains centered on fields other than safe AD systems, such as robotics, mobile assistance, or graphical user interfaces. Key challenges remain largely underexplored, including maintaining semantic fidelity under compression, achieving robust multimodal fusion within tight latency budgets, and integrating VLM inference into AD. Addressing these gaps, the proposed REACT framework introduces a lightweight, edge-deployable VLM architecture specifically designed for cooperative trajectory planning and timely collision avoidance in AD with V2X contexts.

\section{Methodology}

The REACT framework provides real-time, proactive interventions by fusing cooperative perception through V2X communication with multimodal sensor inputs via a specialized VLM. The following sections detail the architecture, computational strategies, and workflows involved in REACT, aligned with the illustrated system overview.

\subsection{Framework Overview}

\begin{figure}
    \centering
    \includegraphics[width=0.75\linewidth]{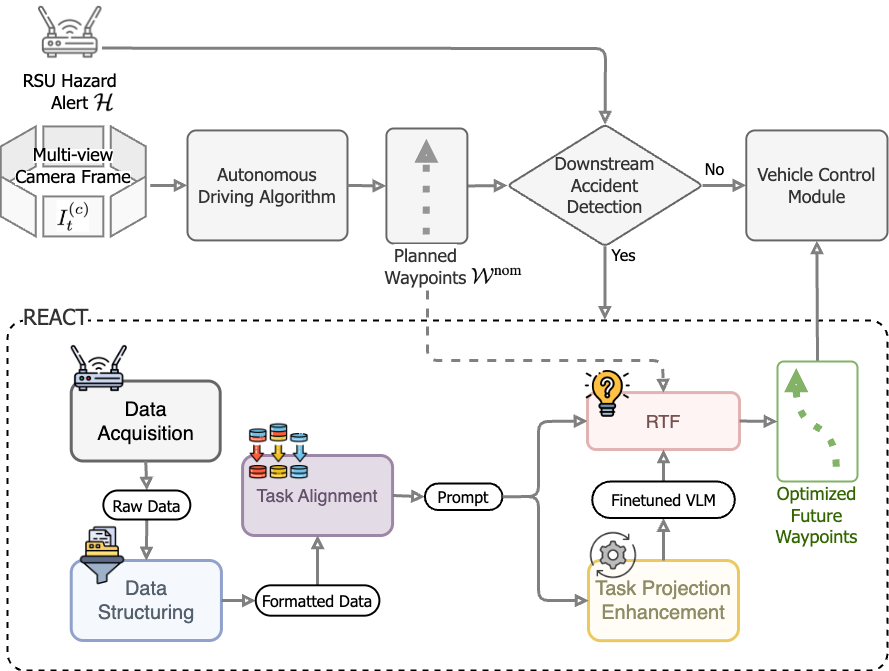}
    \caption{The Overview of the REACT}
    \label{fig:overview}
\end{figure}

The proposed framework builds upon the existing AD system, which generates a nominal trajectory from multi-view camera frames or other onboard perception inputs \cite{chen2024end}. A downstream accident detection module monitors real-time alerts from RSUs and triggers the REACT pipeline only if a hazard is detected. As illustrated in Figure~\ref{fig:overview}, REACT operates in parallel to the AD stack and refines the planned trajectory through five interconnected modules: 

The \textbf{Data Acquisition} module collects raw multimodal inputs, including sensor data from onboard sensors and vehicle systems as well as hazard alerts received via V2X communication from nearby RSUs. These inputs form the foundational data stream for subsequent processing.

The \textbf{Data Structuring} module selects useful information from raw data, filters noise, processes visual data into BEV feature maps, and converts them into structured, model-compatible representations. This ensures the quality and consistency of the data passed to downstream modules.

In the \textbf{Task Alignment} module, the visual and structured textual data is encoded into optimized prompts. These prompts succinctly summarize environmental context, navigation goals, vehicle states, and are specifically designed to maximize the efficiency and accuracy of downstream inference within the VLM.

The \textbf{Task Projection Enhancement} module uses CoT to supervise fine-tuning (SFT) of a lightweight VLM with paired textual and visual inputs. CoT-based SFT improves the model’s ability to interpret prompts by learning intermediate reasoning steps, leading to clearer task understanding, safer predictions, and better generalization during inference.

Finally, the \textbf{Residual Trajectory Fusion (RTF)} module utilizes the fine-tuned VLM to infer the trajectory adjustments and integrates them into the vehicle’s nominal trajectory. The result is an optimized, risk-aware path that enhances safety by proactively avoiding potential collisions.

\begin{figure}
    \centering
    \includegraphics[width=1\linewidth]{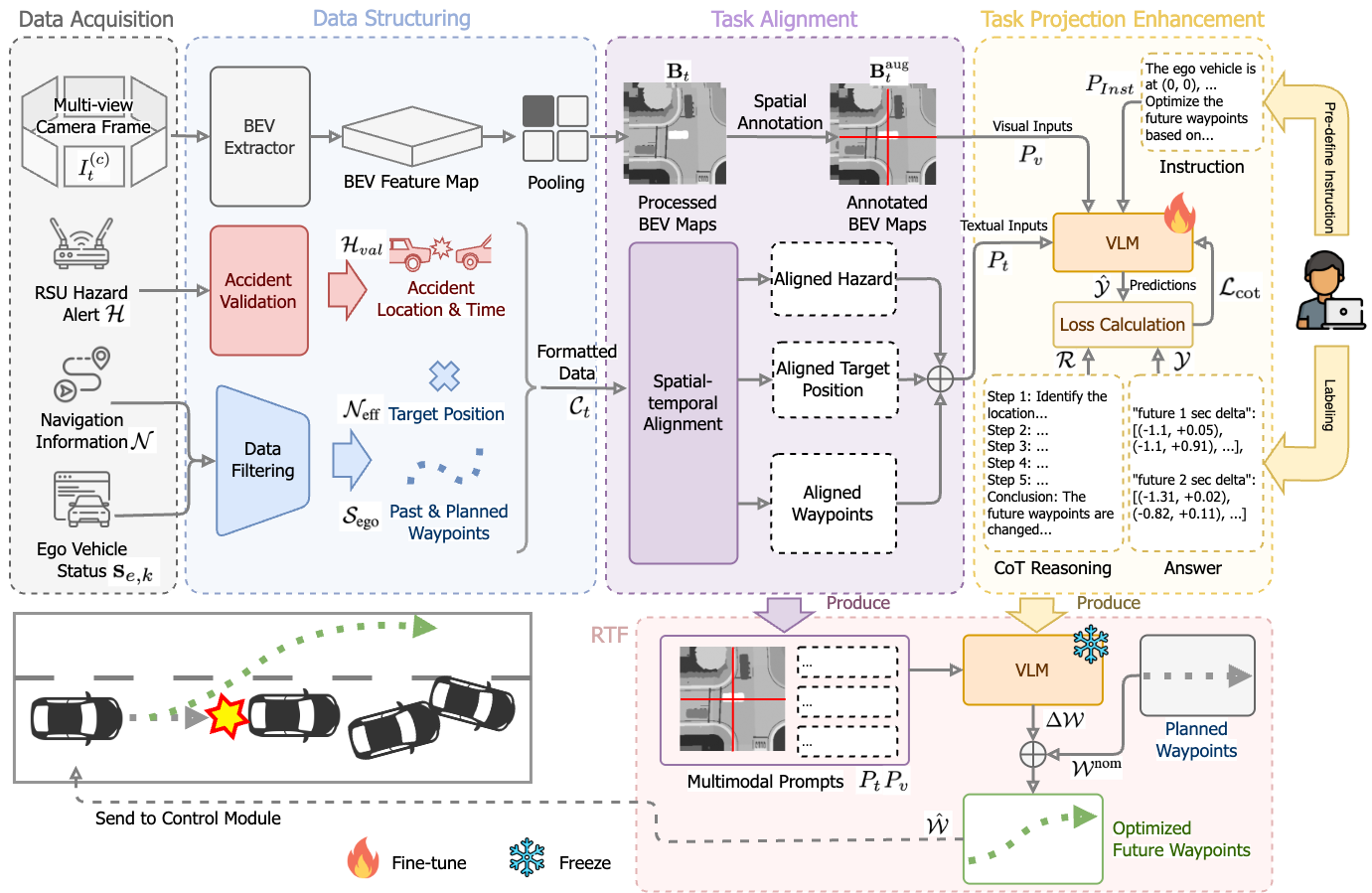}
    \caption{The Architecture of the REACT}
    \label{fig:archi}
\end{figure}

Each module ensures efficient real-time performance suitable for edge-device deployment, collectively enhancing transportation safety through proactive and intelligent trajectory planning. The detailed functionalities of each module are described in the following sections, after the formal configuration of the system. The detailed architecture of REACT is shown in Figure~\ref{fig:archi}.

\subsection{System Configuration}

\begin{table}[htbp]
    \centering
    \caption{Summary of Mathematical Symbols}
    \label{tab:math_symbols}
    \begin{tabular}{|c|l|p{8cm}|}
        \hline
        \textbf{Symbol} & \textbf{Definition} & \textbf{Explanation} \\
        \hline
        $F_{R}$ & RSU coordinate frame & Common spatial reference frame centered at the RSU. \\
        $\hat{\mathbf{x}}_\text{up}$ & RSU x-axis & Unit vector pointing longitudinally in the RSU frame. \\
        $\hat{\mathbf{y}}_\text{right}$ & RSU y-axis & Unit vector pointing laterally (rightward) in the RSU frame. \\
        $\hat{\mathbf{z}}_\text{up}$ & RSU z-axis & Unit vector pointing upward in the RSU frame. \\
        $i$ & Vehicle index & Index for each vehicle in the vehicle set $\mathcal{I}$. \\
        $\mathcal{I}$ & Vehicle set & Set of all indexed vehicles in the scenario. \\
        $k$ & Timestep index & Discrete time index within the time range $\mathcal{K}$. \\
        $\mathcal{K}$ & Time range & Set of all discrete time indices. \\
        $\mathbf{s}_{i,k}$ & Vehicle state vector & State of vehicle $i$ at time $k$, containing position, velocity, and heading. \\
        $x_{i,k}, y_{i,k}, z_{i,k}$ & Vehicle position & Longitudinal, lateral, and vertical position relative to RSU. \\
        $v_{x,i,k}, v_{y,i,k}$ & Vehicle velocity & Velocities in x and y directions within the RSU frame. \\
        $\psi_{i,k}$ & Yaw angle & Heading angle of the vehicle relative to the longitudinal axis. \\
        $\Delta t$ & Time interval & Sampling duration between discrete timesteps. \\
        $r_{i,k}$ & Yaw rate & Rate of change of vehicle $i$’s yaw angle at timestep $k$. \\
        $\mathcal{H}$ & Hazard information & Set of values describing the location and time of a collision. \\
        $x_h, y_h, z_h$ & Hazard position & Location of the hazard relative to the RSU. \\
        $t_h$ & Hazard timestamp & Time to or since the hazard occurred. \\
        $\mathcal{N}$ & Navigation set & Ordered list of waypoints to follow. \\
        $\mathbf{g}_j$ & Waypoint vector & Target position for waypoint $j$ in 2D space. \\
        $x^{\mathrm{ref}}_j, y^{\mathrm{ref}}_j$ & Waypoint coordinates & Reference x-y positions of each waypoint. \\
        $\Delta \mathbf{g}_j$ & Waypoint residual & Displacement vector output from VLM for waypoint $j$. \\
        $\mathcal{W}_{\text{nom}}$ & Nominal waypoints & Planned trajectory before adjustment. \\
 $\Delta \mathcal{W}$ & Waypoint residuals &Full sequence of spatial deltas to adjust $\mathcal{W}_{\text{nom}}$. \\
        $\hat{\mathcal{W}}$ & Optimized waypoints & Refined trajectory after RTF adjustment. \\
        $\mathcal{J}$ & Waypoint index set & Index set $\{1, \dots, M\}$ for all waypoints. \\
        $M$ & Number of waypoints & Total number of waypoints in the navigation set. \\
        $\hat{\mathbf{p}}^{(t)}_{\text{ego}}$ & Ego position at time $t$ & Model-predicted 2D location of the ego vehicle at time $t$. \\
        $\hat{\mathbf{p}}^{(t)}_j$ & Surrounding vehicle $j$'s position & Predicted position of non-ego vehicle $j$ at time $t$. \\
        $\mathcal{P}^{(t)}_{\text{sur}}$ & Surrounding positions at $t$ & Set of all predicted vehicle positions except ego. \\
        ${P}_{Inst}$ & Predefined Instruction & The predefined instructing prompt for VLM reasoning. \\
        $d_{\min}^{(t)}$ & Min distance at time $t$ & Closest gap between ego and any agent at time $t$. \\
        $T$ & Planning horizon timesteps & Total number of predicted steps in the planning horizon. \\
        \hline
    \end{tabular}
\end{table}

To provide a clear foundation for evaluating and validating the proposed REACT framework within typical V2X safety scenarios, this paper defines the problem as follows. This research addresses a critical roadway scenario where an autonomous ego vehicle travels downstream but faces an occluded view due to preceding traffic, preventing direct observation of a collision located ahead. Table~\ref{tab:math_symbols} summarizes the definitions and explanations for each symbol used in this paper.

To provide a unified spatial reference, all vehicle states are defined within a common coordinate system centered at the roadside unit (RSU), denoted as $F_{R}=\{\hat{\mathbf{x}}_\text{up},\,\hat{\mathbf{y}}_\text{right},\,\hat{\mathbf{z}}_\text{up}\}$. In this frame, each vehicle $i \in \mathcal{I}$ in vehicle set $ \mathcal{I} $ at discrete timestep $k \in \mathcal{K}$ in time range $ \mathcal{K} $ is characterized by Eq.~\ref{eq:states}, where $(x,y,z)$ represents the longitudinal, lateral, and vertical positions relative to the RSU origin, $(v_x,v_y)$ denotes velocities in the RSU reference plane, and $\psi$ is the yaw angle representing the vehicle's heading relative to the longitudinal axis.
\begin{equation}
    \label{eq:states}
    \mathbf{s}_{i,k}=[x_{i,k},\,y_{i,k},\,z_{i,k},\,v_{x,i,k},\,v_{y,i,k},\,\psi_{i,k}]^\top,
\end{equation}

Vehicle states evolve according to discrete-time kinematic equations (Eqs.~\ref{eq:xi},~\ref{eq:yi},~\ref{eq:zi},\ref{eq:psi}) assuming constant velocity and highway grade within each sampling interval $\Delta t$, where $r_{i,k}$ denotes the yaw rate obtained from vehicle onboard measurements or estimated from road curvature information. 
\begin{align}
x_{i,k+1}&=x_{i,k}+v_{x,i,k}\,\Delta t, \label{eq:xi}\\
y_{i,k+1}&=y_{i,k}+v_{y,i,k}\,\Delta t, \label{eq:yi}\\
z_{i,k+1}&=z_{i,k}, \label{eq:zi}\\
\psi_{i,k+1}&=\psi_{i,k}+r_{i,k}\,\Delta t, \label{eq:psi}
\end{align}

The RSU broadcasts hazard information as a concise alert in Eq.~\ref{eq:hazard} \cite{zheng2024privacy, he2023towards}, indicating the position $(x_h,y_h,z_h)$ of the collision relative to the RSU and its predicted occurrence timestamp $t_h\geq 0$, where $t_h = 0$ denotes an accident that has already occurred, and $t_h>0$ indicates the accident is anticipated to occur in $t_h$ seconds.
\begin{equation}
    \label{eq:hazard}
    \mathcal{H}=\{x_h,\,y_h,\,z_h,\,t_h\},
\end{equation}

The on-board route planner supplies a finite ordered list of spatial waypoints without fixed timestamps that the ego vehicle must sequentially reach. The navigation set is denoted by Eq.~\ref{eq:nav}, where $\mathcal{J}=\{1,\dots,M\}$ indexes the $M$ waypoints and each vector 
$\mathbf{g}_{j}=(x^{\mathrm{ref}}_{j},\,y^{\mathrm{ref}}_{j},\,z^{\mathrm{ref}}_{j})^{\!\top}$  
specifies the desired longitudinal, lateral, and vertical coordinates of waypoint~$j$ in the RSU‐centred frame $F_{R}$. The final element $\mathbf{g}_{M}$ corresponds to the route destination or the next high-level navigation milestone.  
\begin{equation}
    \label{eq:nav}
    \mathcal{N}\;=\;\bigl\{\,\mathbf{g}_{j}\bigr\}_{j\in\mathcal{J}}
    \;=\;
    \Bigl\{\,\bigl(x^{\mathrm{ref}}_{j},\,y^{\mathrm{ref}}_{j},\,z^{\mathrm{ref}}_{j}\bigr)\Bigr\}_{j=1}^{M},
\end{equation}

\subsection{Data Acquisition}

As the first module of the REACT system shown in Figure~\ref{fig:archi}, the \textit{Data Acquisition} stage collects both visual and non-visual inputs required for downstream reasoning. This module integrates raw sensor data and contextual information from vehicle and infrastructure sources into a unified pipeline. 

Specifically, the ego vehicle $i_{ego}$ captures synchronized multi-view camera frames, denoted by $I^{(c)}t$ at time $t$, where $c$ indexes the camera views. These images are later transformed into a BEV representation to facilitate spatial reasoning. In parallel, the ego vehicle receives hazard alerts $\mathcal{H}$ from the RSU via V2X communication using the C-V2X protocol. Each alert provides the predicted collision location $(x_h, y_h, z_h)$ in the RSU-centered coordinate frame $F_{R}$, along with a timestamp $t_h$ indicating the anticipated hazard time. Complementing this, the onboard route planner supplies a set of $M$ target waypoints $\mathcal{N}$ . Additionally, real-time kinematic data, represented as a state vector $\mathbf{s}_{e,k}$, is obtained through the Controller Area Network (CAN) bus, capturing the ego vehicle’s position, linear velocities, and yaw angle at each discrete timestep $k \in \mathcal{K}$. Together, these raw inputs—$I^{(c)}_t$, $\mathcal{H}$, $\mathcal{N}$, and $\mathbf{s}_{e,k}$—form the basis of the REACT framework’s multimodal reasoning pipeline, enabling the system to perceive beyond the line of sight and anticipate occluded roadway hazards.

\subsection{Data Structuring}

After acquiring raw visual and non-visual data, the \textit{Data Structuring} module transforms heterogeneous inputs into structured representations suitable for downstream reasoning. This process consists of three parallel pipelines: visual processing, accident information validation, and ego vehicle intention \& status extraction.

\textit{Visual Processing}

The multi-view images $I^{(c)}$ are converted into two BEV feature maps to expose short-horizon motion, a current BEV at time $t$ and a past BEV from a short time interval earlier $t{-}\Delta t$, both obtained via 3D reconstruction \cite{mao20233d} and spatial pooling, shown as Eq.~\ref{eq:bev_pair}. The two-frame BEV input summarizes road geometry and all road users at $t{-}\Delta t$ and $t$, providing implicit motion cues over a short temporal window of length $\Delta t$ for downstream reasoning. This dual-frame encoding enhances the model’s ability to understand dynamic contexts without requiring explicit motion vectors.
\begin{equation}
    \label{eq:bev_pair}
    \mathbf{B}^{\mathrm{now}}_{t}=\mathrm{Pool}\ \bigl(\mathcal{F}_{\mathrm{bev}}(I^{(c)}_{t})\bigr),
    \qquad
    \mathbf{B}^{\mathrm{past}}_{t-\Delta t}=\mathrm{Pool}\ \bigl(\mathcal{F}_{\mathrm{bev}}(I^{(c)}_{t-\Delta t})\bigr).
\end{equation}

\textit{Accident Information Validation}

To ensure the reliability of RSU-reported accident information $\mathcal{H}$, a simple rule-based validation step is applied before trajectory adjustment. This module filters out outdated or irrelevant alerts by verifying two primary conditions shown as Eq.~\ref{eq:valid1} and Eq.~\ref{eq:valid2}.
\begin{equation}
    \label{eq:valid1}
    \text{Valid}(\mathcal{H}) = \mathbbm{1}\left[ \mathbf{g}_M \cdot \hat{\mathbf{x}}_{\text{up}} > x_h \;\wedge\; |t_h - t_{\text{now}}| < \delta_t \right]
\end{equation}

\begin{equation}
    \label{eq:valid2}
    \mathcal{H}_{val} = \text{Valid}(\mathcal{H})\cdot\mathcal{H}
\end{equation}

Here, $\mathbf{g}_M \cdot \hat{\mathbf{x}}_{\text{up}} > x_h$ ensures the hazard is located downstream of the final waypoint $\mathbf{g}_M$, while $|t_h - t_{\text{now}}| < \delta_t$ guarantees that the hazard timestamp is recent, bounded by threshold $\delta_t$. If both conditions are satisfied, the accident information is passed to the planning module; otherwise, it is discarded to prevent false positives from outdated or misplaced alerts.

\textit{Ego Intention \& Status Extraction}

To enhance the relevance and clarity of the input prompts for the VLM, and ensure the VLM focuses on the most informative and temporally relevant data for downstream planning, both the navigation goal sequence and the ego status history must be temporally filtered. Specifically, only a short horizon of future waypoints from the navigation set $\mathcal{N}$ is retained, while past waypoints and those far beyond the current context are discarded to reduce noise, shown as Eq.~\ref{eq:navi_filter}.
\begin{equation}
\label{eq:navi_filter}
\mathcal{N}_{\text{eff}} = \left\{ \mathbf{g}_j \in \mathcal{N} \;\middle|\; j \in \{k, k+1, \dots, k+M'\},\; M' \ll M \right\}
\end{equation}

Similarly, for ego motion state $\mathbf{s}_{e,k}$, only a recent history of the past $K$ seconds is preserved to encode short-term dynamics without overwhelming the VLM with redundant temporal details, shown as Eq.~\ref{eq:state_filter}. This paper empirically sets $K = 2$ seconds, as longer trajectory histories do not significantly contribute to immediate decision-making once high-level intent is captured. 
\begin{equation}
\label{eq:state_filter}
\mathcal{S}_{\text{ego}} = \left\{ \mathbf{s}_{e,k'} \;\middle|\; k' \in \{k - K, \dots, k \} \right\}
\end{equation}

The valid hazard alert, navigation intent, and ego state history are encapsulated into a structured symbolic context package $\mathcal{C}_{t}$, defined as Eq.~\ref{eq:extract}.
\begin{equation}
    \label{eq:extract}
    \mathcal{C}_{t} = \{\mathcal{H}_{val},\, \mathcal{N}_{\text{eff}},\, \mathcal{S}_{\text{ego}} \}
\end{equation}

This compact and curated representation serves as the input to the \textit{Task Alignment module}, ensuring that all symbolic inputs are contextually relevant and lightweight for efficient reasoning.

\subsection{Task Alignment}

Effective reasoning with the VLM requires converting the contextual input $\mathcal{C}_t$ into a structured visual-textual prompt that accurately represents the current driving situation. This transformation organizes diverse inputs into a coherent representation. By preserving essential spatial and temporal semantics, the structured prompt enables the VLM to infer context-aware trajectory modifications grounded in both present observations and recent motion trends.

\textit{BEV Coordinate Scaling}

To furnish the VLM with explicit spatial grounding, real‑world coordinates are transferred and overlaid onto the BEV feature map \(\mathbf{B}_{t}\in\mathbb{R}^{C\times H\times W}\), which is defined on a discrete image grid \((u,v)\in\mathcal{G}\subset\{0,\ldots,W{-}1\}\times\{0,\ldots,H{-}1\}\), where \((u,v)\) denotes a pixel in the BEV map and \((u_{0},v_{0})\) is the ego‑vehicle pixel (the anchor). Let \(r\) be the metre‑per‑pixel scale. The ego‑centred coordinates \((x, y\) corresponding to any BEV pixel are obtained by Eq.~\ref{eq:pix2metric_ego}.
\begin{equation}
    \label{eq:pix2metric_ego}
    \begin{bmatrix}x\\[2pt]y\end{bmatrix}
    \;=\;
    r\,
    \begin{bmatrix}u-u_{0}\\[2pt]v-v_{0}\end{bmatrix},
    \qquad (u,v)\in\mathcal{G}.
\end{equation}

For \(\mathbf{B}_{s}\) at time \(s\in\{t,\,t{-}\Delta t\}\), a rendering operator \(\mathcal{R}_{\mathrm{axis}}\) draws short orthogonal strokes from \((u_{0},v_{0})\) with lengths \(H,W\) in pixels and tick spacing \(\kappa\). Denoting by \(\Omega_{s}\subset\mathcal{G}\) the stroked pixel set, the overlaid BEV \(\tilde{\mathbf{B}}_{s}\) is processed by Eq.~\ref{eq:axis_overlay_compact}, where \(\mathcal{S}\) is a simple stroke. 
\begin{equation}
    \label{eq:axis_overlay_compact}
    \tilde{\mathbf{B}}_{s}(u,v)=
    \begin{cases}
        \mathbf{B}_{s}(u,v), & (u,v)\notin \Omega_{s},\\[3pt]
        \mathcal{S}\!\left(\mathbf{B}_{s}(u,v);\;u_{0},v_{0},L_{x},L_{y},\kappa\right), & (u,v)\in \Omega_{s},
    \end{cases}
\end{equation}

The same ego‑anchored overlay is applied to both the current BEV \(\mathbf{B}_{t}\) and the past BEV \(\mathbf{B}_{t{-}\Delta t}\), providing a clear, consistent metric reference while preserving the original channel count.

\textit{Spatial–Temporal Alignment}

To ensure consistent spatial grounding for downstream VLM reasoning, all context elements in $\mathcal{C}_t$ are represented in the ego–centered frame. Since each BEV is already ego–referenced, the hazard set $\mathcal{H}_t$, the effective navigation waypoints $\mathcal{N}_{\text{eff}}$, and the ego trajectory history $\mathcal{S}_{\text{ego}}$ are translated into the same frame by Eq.~\ref{eq:spat_align_all_rev}, where $(x,y)$ denotes any global position from $\mathcal{H}_t$, $\mathcal{N}_{\text{eff}}$, or $\mathcal{S}_{\text{ego}}$, and $(x_{e,k},y_{e,k})$ is the current ego position.
\begin{equation}
    \label{eq:spat_align_all_rev}
    (x^{\mathrm{rel}},\,y^{\mathrm{rel}}) \;=\; (x,\,y) - (x_{e,k},\,y_{e,k}),
\end{equation}

Heterogeneous timestamps are normalized to a common zero–centered timeline to support causal reasoning, shown as Eq.~\ref{eq:temp_align_rev}, where \(t_{\text{0}}\) denotes the present.
\begin{equation}
    \label{eq:temp_align_rev}
    t' \;=\; t - t_{\text{0}}
\end{equation}

The overall alignment process for prompt generation is then expressed as Eq.~\ref{eq:tprompt_aligned_rev}, while the visual prompt consumes the \(\mathbf{B}_{t_0}\) and \(\mathbf{B}_{t_0{-}\Delta t}\) as Eq.~\ref{eq:vprompt_aligned_rev}.
\begin{equation}
    \label{eq:tprompt_aligned_rev}
    P_{t} \;=\; \mathcal{E}_{\mathrm{prompt}}\!\left(\,
    \mathcal{A}_{\mathrm{spat}}\!\bigl(\mathcal{A}_{\mathrm{temp}}(\mathcal{C}_t)\bigr)\,
    \right),
\end{equation}
\begin{equation}
    \label{eq:vprompt_aligned_rev}
    P_{v} \;=\; \mathcal{E}_{\mathrm{prompt}}\!\left(\,
    \tilde{\mathbf{B}}_{t_0},\; \tilde{\mathbf{B}}_{\,t_0-\Delta t}
    \right).
\end{equation}
Here, $\mathcal{A}_{\mathrm{temp}}$ applies Eq.~\eqref{eq:temp_align_rev}, $\mathcal{A}_{\mathrm{spat}}$ applies Eq.~\eqref{eq:spat_align_all_rev}, and $\mathcal{E}_{\mathrm{prompt}}(\cdot)$ subsequently encodes these aligned inputs into structured
textual and visual prompts \(P_{t}\) and \(P_{v}\) , which enhance the VLM’s understanding of dynamic driving contexts for safe and interpretable motion planning.

\subsection{Task Projection Enhancement}

To improve reasoning accuracy and task alignment, REACT applies CoT SFT, which teaches the VLM to follow structured logic before generating predictions. This enhances interpretability, spatial awareness, and risk-sensitive planning—enabling the model to generalize better under complex scenarios while remaining efficient on edge devices.

\textit{Instruction and Input Generation}

For each training instance, the training set is prepared that includes: (1) a predefined instruction $\mathcal{P}_{Inst}$ explaining the optimization goal, (2) a multimodal prompt $P = \{P_t, P_v\}$ representing the scenario context, (3) a human-written CoT-style reasoning trace $\mathcal{R}$ that describes the step-by-step planning rationale, and (4) a human-labeled answer. Together, they form the input set as Eq.~\ref{eq:cot_input}, where $\|$ denotes concatenation across instruction, symbolic, visual, and CoT components.
\begin{equation}
    \label{eq:cot_input}
    \mathcal{X} = \bigl[{P}_{Inst}\, \|\, P_t\, \|\, P_v\, \|\, \mathcal{R} \bigr]
\end{equation}

\textit{Ground-Truth Output Format}

The target output $\mathcal{Y}$ consists of delta displacements for future waypoints shown as Eq.~\ref{eq:gt_output}, where each $\Delta \mathcal{W}^*_j$ represents the ground truth adjustment applied to waypoint $j$ in the navigation plan.
\begin{equation}
    \label{eq:gt_output}
    \mathcal{Y} = \left\{\Delta \mathcal{W}^*_j = \left(\Delta x^*_j, \Delta y^*_j\right)\right\}_{j=1}^M
\end{equation}

\textit{Training Objective}

During training, the VLM takes $\mathcal{X}$ as input and generates predicted outputs $\hat{\mathcal{Y}}$. Let $\hat{\mathcal{Y}} = \{\hat{y}_1, \hat{y}_2, \dots, \hat{y}_L\}$ denote the VLM’s token predictions corresponding to the output tokens in $\mathcal{Y}$. The model is trained using a causal cross-entropy loss, which encourages the VLM to autoregressively predict the reasoning steps and final outputs conditioned on the full prompt, shown as Eq.~\ref{eq:causal_loss}.
\begin{equation}
    \label{eq:causal_loss}
    \mathcal{L}_{\mathrm{cot}} = - \sum_{\ell=1}^{L} \log p\left(\hat{y}_\ell \mid \hat{y}_{<\ell},\, \mathcal{X} \right)
\end{equation}

\subsection{RTF}

To balance processing latency and output quality, REACT predicts only trajectory adjustments rather than generating complete future trajectories. The RTF module refines the nominal trajectory $\mathcal{W}_{\text{nom}}$ using residual corrections derived from multimodal VLM outputs. Specifically, the RTF process decodes VLM-predicted residuals $\Delta \mathcal{W}$ and applies them to the existing trajectory, enabling proactive modifications without discarding the original plan. Compared to direct trajectory generation, RTF reduces inference latency and output token complexity while preserving motion continuity.

Given the multimodal prompts $P_t$ and $P_v$ generated during task alignment, the VLM outputs a sequence of waypoint-wise residual displacements, shown as Eq.~\ref{eq:delta_waypoints}, where each residual $\Delta \mathbf{g}_j = (\delta x_j,\, \delta y_j)$ denotes the spatial offset for waypoint $\mathbf{g}_j \in \mathcal{W}_{\text{nom}}$.
\begin{equation}
    \label{eq:delta_waypoints}
    \Delta \mathcal{W} = \{\Delta \mathbf{g}_1,\, \Delta \mathbf{g}_2,\, \dots,\, \Delta \mathbf{g}_M\}
\end{equation}

The final optimized trajectory $\hat{\mathcal{W}}$ is computed by element-wise addition in Eq.\ref{eq:waypoint_update}.
\begin{equation}
    \label{eq:waypoint_update}
    \hat{\mathcal{W}} = \{ \mathbf{g}_j + \Delta \mathbf{g}_j \;\;|\;\; \mathbf{g}_j \in \mathcal{W}_{\text{nom}} \}
\end{equation}

This approach ensures that the updated plan $\hat{\mathcal{W}}$ remains closely aligned with the original nominal trajectory while adaptively refining it to avoid predicted risks. The optimized trajectory is subsequently provided to the vehicle control module for execution.

\section{Edge Computing Adaptation}

Deploying REACT on resource-constrained edge devices presents unique challenges, including limited memory bandwidth and stringent latency constraints. To address these challenges, this paper proposes a series of tailored optimization strategies, including model quantization, enhanced attention computation, and textual and visual token reductions.

\subsection{Adaptation Overview}

\begin{figure}
    \centering
    \includegraphics[width=0.5\linewidth]{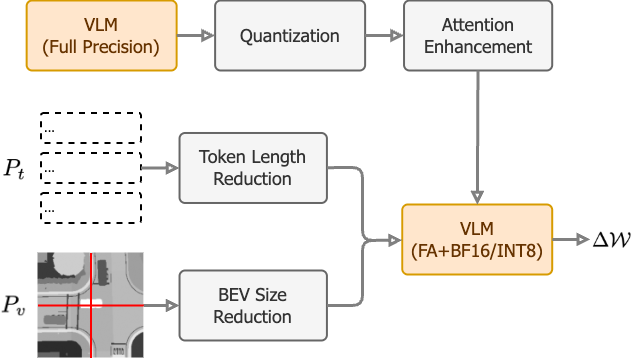}
    \caption{Edge computing adaption process}
    \label{fig:edge_adaption}
\end{figure}

Figure~\ref{fig:edge_adaption} illustrates the overall optimization pipeline for deploying REACT on edge devices. The adaptation process begins with the selection of an appropriate VLM backbone. SmolVLM2-500M is chosen due to its balance between compact model size and sufficient capability to accurately follow structured prompts. Larger models, like Qwen\_2.5VL (3B and 7B) and LLaVA\_1.5 (7B), exhibited prohibitively high latency, while smaller alternatives, such as SmolVLM2-256M, failed to consistently adhere to the required output format. Following model selection, this study performs quantization, converting the model weights from full precision (32-bit) to BF16 (16-bit) or INT8 (8-bit) precision to reduce memory usage and enhance inference efficiency. Next, the "Attention Enhancement" step incorporates scaled dot-product attention improvements, optimizing transformer self-attention calculations and further reducing computational overhead. To minimize inference latency, additional reductions in input complexity are implemented. In the "Token Length Reduction" step, the textual prompts are reduced in token length further to reduce the computation cost. Concurrently, the "BEV Size Reduction" step compresses the BEV feature map resolution, substantially decreasing the number of visual tokens required for inference.

\subsection{Model Quantization}

Model quantization involves converting the model weights from full 32-bit floating-point precision to lower bit-depth precision such as INT8 or FP16. This method reduces memory usage and computational demands by lowering the numerical precision required for storage and computation. Consequently, quantization enables faster inference and decreased power consumption, making real-time deployment feasible on edge devices without critically compromising model accuracy.

\subsection{Attention Enhancement}

The attention enhancement strategy utilizes scaled dot-product attention (SDPA) to optimize the transformer's self-attention mechanism. By simplifying and streamlining matrix multiplication operations inherent in attention calculations, SDPA substantially decreases computational complexity. This approach improves the efficiency of inference operations, thereby enabling quicker and more responsive real-time predictions within computationally restricted environments.

\subsection{Token Length Reduction}

Token length reduction optimizes the textual input by strategically shortening the length of prompts provided to the model. This involves rounding numerical values, removing redundant or non-essential textual elements, and retaining only the critical semantic information necessary for contextually accurate predictions. By reducing input complexity, this method directly decreases inference latency, allowing for faster processing without diminishing essential contextual reasoning capabilities.

\subsection{BEV Size Reduction}

The BEV size reduction strategy compresses the spatial resolution of the BEV feature maps through pooling operations, thereby decreasing the number of visual tokens required during model inference. Reducing visual token count minimizes computational load and memory bandwidth demands, facilitating more efficient inference. This enables REACT to maintain real-time responsiveness and performance standards crucial for edge-based autonomous driving applications.

\section{Experiment}

\subsection{Experimental Setup}

This study utilizes the \textit{DeepAccident} dataset \cite{wang2024deepaccident}, which comprises 691 accident scenarios simulated in CARLA \cite{Dosovitskiy17}, grounded in real-world crash reports. The dataset provides comprehensive coverage of safety-critical driving contexts by integrating diverse weather and lighting conditions, varied traffic densities, and both crash and non-crash outcomes. It includes six high-risk intersection scenarios spanning signalized and unsignalized environments, and is evenly distributed across diverse geographical areas such as small town, downtown apartment and commercial districts, rural roads, and highways. Each scenario offers multi-view camera and LiDAR data from vehicles and an RSU, supporting motion prediction and collision reasoning tasks, and enabling robust training of V2X-based collision avoidance models.

To augment the training set, this paper selects 10 to 20 frames, depending on scenario length, immediately before and after each collision or near-miss event. For every selected frame, vehicle positions are extracted over a 2-second window preceding the frame and a 2-second window following it, together forming a comprehensive data sample. Navigation targets are assigned based on the ego vehicle’s position in the final frame. For scenarios that terminate at the collision frame, the target position is manually specified. After preprocessing, the resulting training set consists of 5,959 valid samples for model fine-tuning. To maintain compatibility with real-time inference constraints on edge devices, BEV feature maps are compressed per frame to ensure the total input token count remains below 1,600. The validation set is preprocessed in the same way as the training set and contains 942 samples for model evaluation.

To evaluate deployment performance on edge platforms, the fine-tuned models are tested on two representative devices: the Jetson Orin Nano (Nano) and the Jetson AGX Xavier (AGX). However, due to memory constraints causing significant latency increases on the Nano, subsequent performance metrics presented in this study focus exclusively on the AGX platform.

Hyperparameter tuning identifies a learning rate of $1 \times 10^{-3}$ as optimal for training stability and convergence. Fine-tuning is conducted with a batch size of 6, a gradient accumulation step of 4, 25 warmup steps, and 20 epochs, resulting in a total of 4,956 steps across the complete training set.

This study evaluates various edge adaptation strategies, as detailed in the ablation studies. Based on the experimental outcomes, the optimal configuration utilizes 16-bit quantization, a token length of 1500, and a BEV resolution of $64 \times 64$. Subsequent experiments adopt this optimized configuration to ensure optimal performance.

\subsection{Evaluation Metrics}

Following the evaluation protocol in \cite{wang2024deepaccident}, this study assesses model performance using four complementary metrics: Video Panoptic Quality (VPQ) \cite{hu2021fiery}, Mean Intersection-over-Union (mIOU), Minimum Average Displacement Error (minADE), and Minimum Final Displacement Error (minFDE).

\textbf{Video Panoptic Quality (VPQ)}

VPQ quantifies the pixel-level accuracy of predicted motion maps by measuring how well the model forecasts future grid-cell occupancy. Let $\hat{M}t$ and $M_t$ denote the predicted and ground-truth motion maps at time step $t$, respectively. VPQ is defined as Eq.~\ref{eq:vpq}, where the intersection and union are computed over all spatial cells in the BEV grid.
\begin{equation}
    \label{eq:vpq}
    \text{VPQ} = \frac{1}{T} \sum_{t=1}^{T} \frac{\hat{M}_t \cap M_t}{\hat{M}_t \cup M_t},
\end{equation}

\textbf{Mean Intersection-over-Union (mIOU)}

MIOU evaluates the region-wise overlap between predicted and ground-truth occupied areas across all frames and accident scenarios. For each scenario $s$, the IOU is computed as Eq.~\ref{eq:iou}, where $\hat{A}_s$ and $A_s$ represent predicted and ground-truth occupancy regions, respectively.
\begin{equation}
    \label{eq:iou}
    \text{IOU}_s = \frac{|\hat{A}_s \cap A_s|}{|\hat{A}_s \cup A_s|},
\end{equation}
The overall MIOU is calculated as Eq.~\ref{eq:miou} with $S$ denoting the number of accident scenarios.
\begin{equation}
    \label{eq:miou}
    \text{mIOU} = \frac{1}{S} \sum_{s=1}^{S} \text{IOU}_s.
\end{equation}

\textbf{Minimum Average Displacement Error (minADE)}

MinADE measures the mean Euclidean distance between predicted and ground-truth waypoints over a prediction horizon. Given $N$ candidate trajectories $\{ \hat{\mathcal{W}}_n \}_{n=1}^{N}$ and ground-truth trajectory $\mathcal{W}^{*} = \{ \mathbf{w}^*_1, \dots, \mathbf{w}^*_T \}$, the ADE for candidate $n$ is calculated as Eq.~\ref{eq:ade} and minADE is defined as Eq.~\ref{eq:minade}.
\begin{equation}
    \label{eq:ade}
    \text{ADE}_n = \frac{1}{T} \sum_{t=1}^{T} \left\| \hat{\mathbf{w}}_{n,t} - \mathbf{w}^*_t \right\|_2,
\end{equation}
\begin{equation}
    \label{eq:minade}
    \text{minADE} = \min_{n \in \{1,\dots,N\}} \text{ADE}_n.
\end{equation}

\textbf{Minimum Final Displacement Error (minFDE)}

minFDE captures the final-point accuracy of trajectory predictions by computing the Euclidean distance between each predicted endpoint and the ground-truth destination, which is calculated as Eq.~\ref{eq:minfde}.
\begin{equation}
    \label{eq:minfde}
    \text{minFDE} = \min_{n \in \{1,\dots,N\}} \left\| \hat{\mathbf{w}}_{n,T} - \mathbf{w}^*_T \right\|_2.
\end{equation}
Together, these metrics provide a comprehensive evaluation of spatial alignment (VPQ, mIOU) and trajectory accuracy (minADE, minFDE), facilitating consistent and rigorous benchmarking against prior work.

\textbf{Collision Rate Reduction (CRR)}

To measure how effectively REACT reduces collision risk compared to a baseline AD model, this study calculates a frame-wise collision rate across accident-related frames \(f \in \{1,\dots,N\}\). For each frame \(f\), a collision is determined by computing the minimum distance between the predicted ego vehicle position \(\hat{\mathbf{p}}^{(f)}_{\text{ego}}\) and all predicted surrounding vehicle positions \(\{\hat{\mathbf{p}}^{(f)}_j\}_{j\in\mathcal{I}^{(f)}_{\text{sur}}}\), where \(\mathcal{I}^{(f)}_{\text{sur}}\) indexes non-ego agents. Define the set of surrounding positions in Eq.~\ref{eq:set_pos}, and the frame-wise minimum separation distance is then calculated by Eq.~\ref{eq:ego_min_sep}.
\begin{equation}
    \label{eq:set_pos}
    \mathcal{P}^{(f)}_{\text{sur}} \triangleq \bigl\{\hat{\mathbf{p}}^{(f)}_j \;\big|\; j\in\mathcal{I}^{(f)}_{\text{sur}}\bigr\}
\end{equation}
\begin{equation}
    \label{eq:ego_min_sep}
    d_{\min}^{(f)} \;=\; \min_{\mathbf{q}\in \mathcal{P}^{(f)}_{\text{sur}}}
    \bigl\| \hat{\mathbf{p}}^{(f)}_{\text{ego}} - \mathbf{q} \bigr\|_{2}
\end{equation}

A collision is recorded for frame \( f \) if the minimum distance is less than 5 meters in Eq.~\ref{eq:collision2}.
\begin{equation}
    \label{eq:collision2}
    F_{\mathrm{col}}^{(f)} = 
    \begin{cases}
        1, & \text{if } d_{\min}^{(f)} < 5 \text{ m} \\
        0, & \text{otherwise}
    \end{cases}
\end{equation}
The collision rate \( \mathcal{C} \) is then defined as Eq.~\ref{eq:collision3}.
\begin{equation}
    \label{eq:collision3}
    \mathcal{C} = \frac{1}{N} \sum_{f=1}^{N} F_{\mathrm{col}}^{(f)}
\end{equation}

Finally, the collision rate reduction (CRR) achieved by REACT's collision rate \( \mathcal{C}_{\text{REACT}} \) compared to the baseline's rate \( \mathcal{C}_{\text{baseline}} \) is computed as Eq.~\ref{eq:crr}.
\begin{equation}
\label{eq:crr}
\text{CRR} = \frac{\mathcal{C}_{\text{baseline}} - \mathcal{C}_{\text{REACT}}}{\mathcal{C}_{\text{baseline}}}
\end{equation}

\textbf{Minimum Clearance Distance (MCD)}

To evaluate how conservatively the ego vehicle plans its motion relative to surrounding traffic, this paper introduces the Minimum Clearance Distance (MCD) metric, which quantifies the closest proximity between the ego vehicle and any surrounding agent during the future planning horizon. For each future timestep \( t \in \{1, \dots, T\} \), the model-predicted position of the ego vehicle is denoted as \( \hat{\mathbf{p}}^{(t)}_{\text{ego}} \), and the set of predicted positions for all surrounding vehicles is defined as \( \mathcal{P}^{(t)}_{\text{sur}} = \{\hat{\mathbf{p}}^{(t)}_{j} \mid j \in \mathcal{I}_{\text{sur}} \} \), where \( \mathcal{I}_{\text{sur}} \) is the index set of surrounding agents.

The timestep-wise minimum distance is computed by Eq.~\ref{eq:mcd_timestep}.
\begin{equation}
    \label{eq:mcd_timestep}
    d_{\min}^{(t)} =
    \min_{\hat{\mathbf{p}}^{(t)}_{j} \in \mathcal{P}^{(t)}_{\text{sur}}}
    \bigl\lVert
    \hat{\mathbf{p}}^{(t)}_{\text{ego}} - \hat{\mathbf{p}}^{(t)}_{j}
    \bigr\rVert_2.
\end{equation}

The overall MCD across the entire planning horizon is defined as Eq.~\ref{eq:mcd_overall}.
\begin{equation}
    \label{eq:mcd_overall}
    \mathrm{MCD} = \min_{t \in \{1, \dots, T\}} d_{\min}^{(t)}.
\end{equation}

The MCD captures the minimum separation maintained by the ego vehicle from nearby traffic, reflecting the spatial safety margin enforced by the planned trajectory.

\subsection{Model Finetuning}

The fine-tuning process of SmolVLM2-500M within the REACT framework exhibited stable convergence over the course of training. The training loss decreased steadily from above 0.60 to below 0.05 within the first 4,700 steps, indicating effective learning without early overfitting. Simultaneously, the minADE was monitored as a proxy for validation loss and showed a notable reduction of approximately 0.5 meters, with the most rapid improvement occurring between steps 2,000 and 3,000. Although minor fluctuations in minADE were observed, likely due to the diverse driving scenarios in the evaluation set, it remained low and stable after convergence. A slight increase in both training loss and minADE beyond step 4,700 suggested the onset of overfitting. Consequently, the model checkpoint at step 3,400 was selected for deployment, balancing accuracy and generalization.

\subsection{Results}

\begin{table}
    \centering
    \caption{Comparison of Motion Prediction Accuracy with Prior Fusion Methods. ↑: higher is better; bold value means the highest value.}
    \label{tab:motion_comparison}
    \begin{tabular}{>{\centering\arraybackslash}p{0.2\linewidth}>{\centering\arraybackslash}p{0.12\linewidth}>{\centering\arraybackslash}p{0.12\linewidth}>{\centering\arraybackslash}p{0.12\linewidth}>{\centering\arraybackslash}p{0.1\linewidth}>{\centering\arraybackslash}p{0.1\linewidth}}
        \toprule
        \textbf{Method}  &\textbf{Perception Backbones} &\textbf{Fusion Backbones} &\textbf{Prediction Backbones}& \textbf{mIOU ↑} & \textbf{VPQ ↑} \\
        \midrule
            V2XFormer+Average Fusion \cite{wang2024deepaccident}&Transformer&Simple average &Transformer& 52.1 & 39.5 \\
            V2XFormer+DiscoNet \cite{mehr2019disconet}&Transformer&Autoencoder &Transformer& 54.2 & 42.0 \\
            V2XFormer+V2X-ViT \cite{xu2022v2x}&Transformer &Transformer &Transformer & 55.1 & 43.2 \\
            V2XFormer+CoBEVT \cite{xu2022cobevt}&Transformer &Transformer &Transformer & 56.2 & 44.0 \\
            AccidentGPT \cite{wang2023accidentgpt}      & Transformer & Transformer+VLM&LLM& 57.3 & 45.2 \\
        \textbf{REACT (ours)}  & Transformer & VLM&VLM& \textbf{59.1} & \textbf{48.2} \\
        \bottomrule
    \end{tabular}
\end{table}

\begin{figure}
    \centering
    \includegraphics[width=0.75\linewidth]{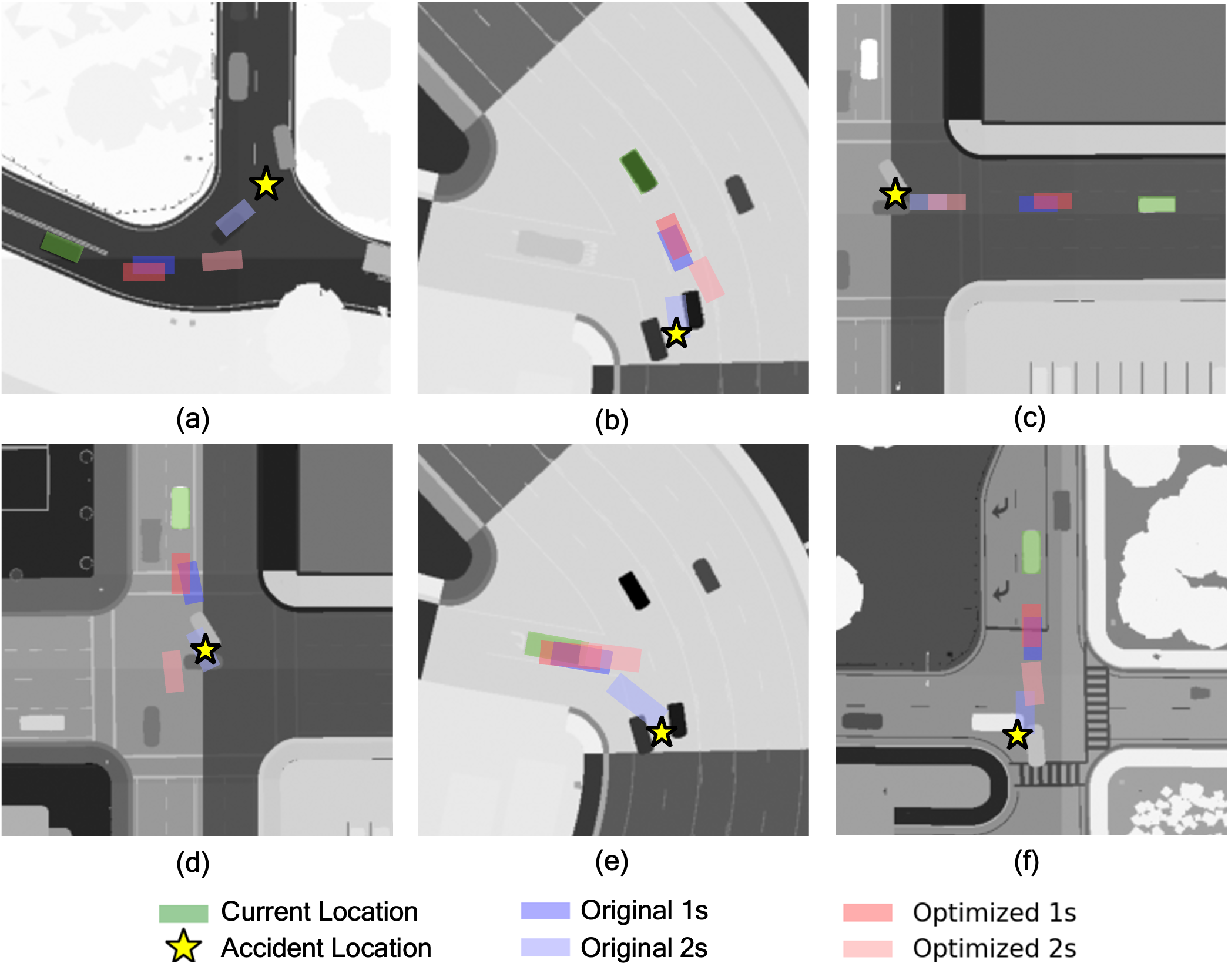}
    \caption{Qualitative simple examples of REACT predictions in accident-prone scenes. (a) Left-turn avoidance at T-intersection, (b) proactive steering on curved multi-lane road, (c) early stop before mid-block collision, (e) conflict avoidance via lateral deviation, (f) curved-road bypass with accident ahead, (g) rear-end prevention through full stop.}
    \label{fig:qualitative_results}
\end{figure}

\begin{figure}
    \centering
    \includegraphics[width=0.75\linewidth]{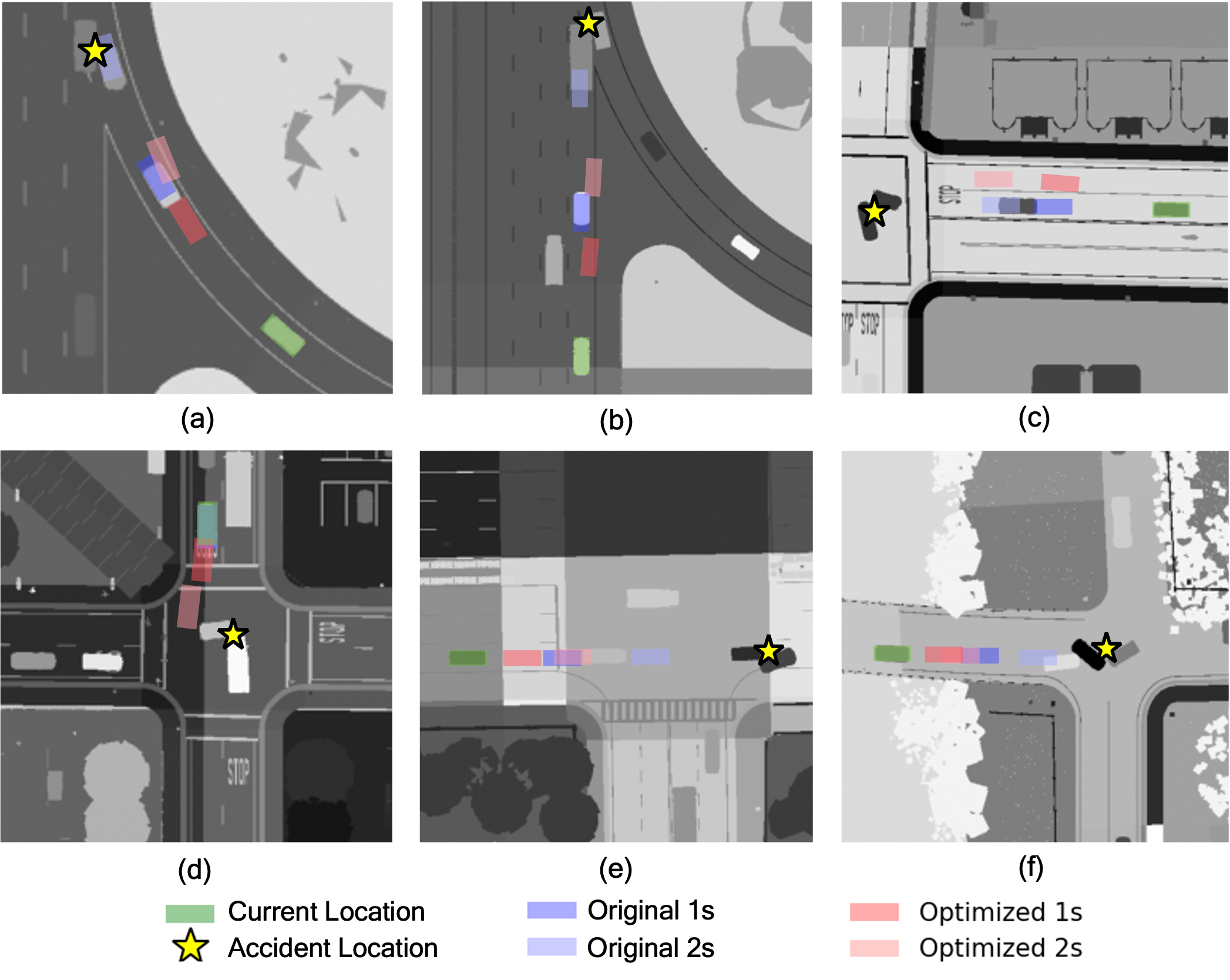}
    \caption{Qualitative complex examples of REACT predictions in accident-prone scenes. (a) Pull right to avoid accident on ramp, (b) pull right to avoid accident on highway, (c) change lane before stop sign to avoid congestion, (d) first come first go at stop sign, (e) stop before entering intersection, (f) early stop to avoid car pile-ups.}
    \label{fig:qualitative_results_complex}
\end{figure}

\subsubsection{Motion Prediction Comparison}

This paper benchmarks REACT against five representative V2X fusion baselines on the DeepAccident dataset, as shown in Table~\ref{tab:motion_comparison}. Average Fusion and DiscoNet utilize early and intermediate fusion schemes, combining multi-agent features at fixed stages, while V2X-ViT and CoBEVT leverage Transformer-based fusion backbones for enhanced spatial reasoning. AccidentGPT introduces language modeling by encoding accident context into prompts and decoding predicted motions with a general-purpose LLM. In contrast, REACT integrates a task-specific VLM that fuses BEV perception with structured symbolic prompts and directly refines trajectory predictions through instruction following. By unifying contextual reasoning and motion planning within a single VLM, REACT achieves superior spatial alignment and prediction accuracy. This alignment between domain-specific knowledge and trajectory generation enables REACT to outperform all baselines, including AccidentGPT, with higher mIOU and a 3-point gain in VPQ using a smaller, more efficient model.

\subsubsection{Qualitative Results}

This research adopts the same motion planning setting as defined in the dataset. To address the missing post-accident trajectory predictions, the BEVerse-tiny model \cite{zhang2022beverse, huang2021bevdet} is selected as the autonomous driving baseline, and its predicted positions are visualized as blue ``Original'' boxes for comparison in Figure~\ref{fig:qualitative_results} and Figure~\ref{fig:qualitative_results_complex}. The \textit{simple scenarios} refers to cases where REACT optimizes the trajectories of vehicles immediately following the crashed vehicles, serving to evaluate the method's direct effectiveness. In contrast, \textit{complex scenarios} involves optimizing vehicles further upstream, where indirect visual cues must be inferred to prevent potential invisible collisions.

Figure~\ref{fig:qualitative_results} illustrates REACT’s qualitative motion predictions across a range of accident-prone \textit{simple scenarios}. In (a), the ego vehicle performs a cautious left-turn adjustment at a T-intersection to avoid a merging conflict with two lead vehicles. In (b), it veers left on a curved multi-lane road to bypass a forward-side collision. In (d), a lateral deviation avoids a crash between the front vehicle and a red-light violator. Scenario (e), a variation of (b), focuses on a different agent, showing the ego vehicle reversing and halting to avoid both the frontal crash and an approaching vehicle from the left. Scenarios (c) and (f) highlight REACT’s early stopping behavior. In (c), the vehicle halts mid-block before reaching a crashed vehicle—unlike the baseline, which fails to avoid it. In (f), REACT prevents a rear-end collision by stopping just before a stationary vehicle affected by side traffic. These examples demonstrate REACT’s ability to proactively alter trajectories in anticipation of downstream hazards.

Figure~\ref{fig:qualitative_results_complex} further illustrates REACT’s advantages in \textit{complex scenarios}. In (a) and (b), REACT performs rightward deviations on highways to avoid both downstream accidents and upstream traffic congestion, effectively utilizing context beyond the ego lane. Scenario (c) shows a proactive lane change before a stop sign to bypass buildup behind a blocked vehicle. In (d), REACT yields at a stop-controlled intersection following a first-come, first-go protocol, in contrast to the planned trajectory that continues to wait unnecessarily, demonstrating socially aware, rule, compliant navigation. Scenarios (e) and (f) underscore REACT’s foresight in initiating early stops. In (e), it halts before entering a crash-prone intersection, improving safety and adhering to traffic regulations that prohibit stopping within intersections. In (f), REACT stops in anticipation of multiple vehicle pile-ups ahead. Together, these examples showcase REACT’s ability to predict and mitigate risks through timely, context-aware trajectory adjustments that comply with traffic rules across diverse urban driving environments.

\subsubsection{REACT Performance under Various Conditions}

\begin{figure}
    \centering
    \includegraphics[width=0.7\linewidth]{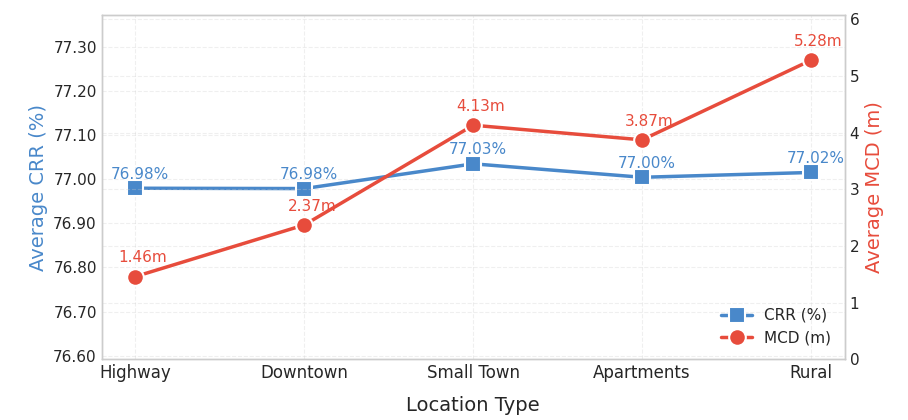}
    \caption{Effect of location type on REACT performance}
    \label{fig:location}
\end{figure}

To evaluate REACT's effectiveness across diverse driving environments, this study calculates collision rates with and without REACT on the validation set to derive the CRR in percentage. It also measures the MCD between the ego vehicle and surrounding objects. Table~\ref{tab:weather} and Figure~\ref{fig:location} summarize the average CRR and corresponding MCD under varying environmental and spatial conditions.

\begin{table}
    \centering
    \caption{REACT performance by weather and time of day}
    \label{tab:weather}
    \begin{tabular}{c|cccc|cccc}
        \toprule
 & &\multicolumn{2}{c}{\textbf{CRR}}&& &\multicolumn{2}{c}{\textbf{MCD}}&\\
        \textbf{Time of Day/Weather} & \textbf{Clear} & \textbf{Cloudy} & \textbf{Rainy} & \textbf{Wet}  & \textbf{Clear} & \textbf{Cloudy} & \textbf{Rainy} &\textbf{Wet}  \\
        \midrule
        \textbf{Noon}    & 77.04\% & 76.96\% & 77.00\%& 76.98\%  & 6.24 & 4.60 & 2.22 &3.32 \\
        \textbf{Sunset}  & 77.07\%& 76.98\% & 77.01\% & 76.95\%& 4.54 & 4.71 & 2.06 &1.66 \\
        \textbf{Night}   & 76.96\% & -- & 77.00\%& --  & 3.08 & -- & 1.09 &-- \\
        \bottomrule
    \end{tabular}
\end{table}

Table~\ref{tab:weather} analyzes REACT’s performance across different combinations of weather and time of day. Across all 11 conditions, the average CRR remains tightly bounded between 76.95\% and 77.07\%, demonstrating REACT’s consistent performance and strong generalization under weather-related perturbations. The ``Clear, Sunset'' condition yields the highest CRR at 77.07\% , suggesting that REACT acts more proactively and maintains greater clearance in optimal conditions. In contrast, ``Wet, Sunset'' exhibits the lowest CRR at 76.95\% with a tighter MCD under 2 meters, possibly due to low light angles and surface reflections affecting perception, combined with slippery roads prompting more conservative behavior. Even in adverse scenarios such as ``Rainy, Night'' and ``Wet, Noon,'' REACT maintains CRR near or above 76\%, indicating resilience to reduced visibility and degraded road conditions. Notably, ``Rainy, Night'' results in the lowest clearance, suggesting REACT responds more cautiously in low-visibility environments. Overall, clearer environmental conditions correlate with higher MCDs, while lower MCDs in rainy or dim settings reflect REACT’s spatial adaptability without compromising safety.

This study further evaluates REACT’s behavior across five distinct location types, as shown in Figure~\ref{fig:location}. ``Small Town'' areas, characterized by low speed limits and minimal vehicle or vulnerable road user (VRU) traffic, exhibit the highest CRR and a moderate MCD, indicating effective and spatially cautious trajectory planning. ``Rural'' roads show the highest MCD and a slightly lower but still strong CRR of 77.01\%, suggesting conservative behavior in open, low-density environments. Conversely, ``Highway'' and ``Downtown'' settings yield the lowest MCDs and the smallest reductions in collision rate. This may be attributed to higher speeds on highways and denser traffic in urban centers, which reduce both available space and reaction time. These conditions challenge evasive trajectory planning, yet REACT still maintains a high CRR, reinforcing its effectiveness even in constrained and dynamic scenarios.

\subsection{Ablation Studies}

This section presents ablation experiments to evaluate the impact of three key design choices in REACT: (1) VLM model size and BEV feature resolution, (2) edge adaptation strategies, and (3) the inclusion of the RTF module.

\begin{figure}
    \centering
    \includegraphics[width=0.75\linewidth]{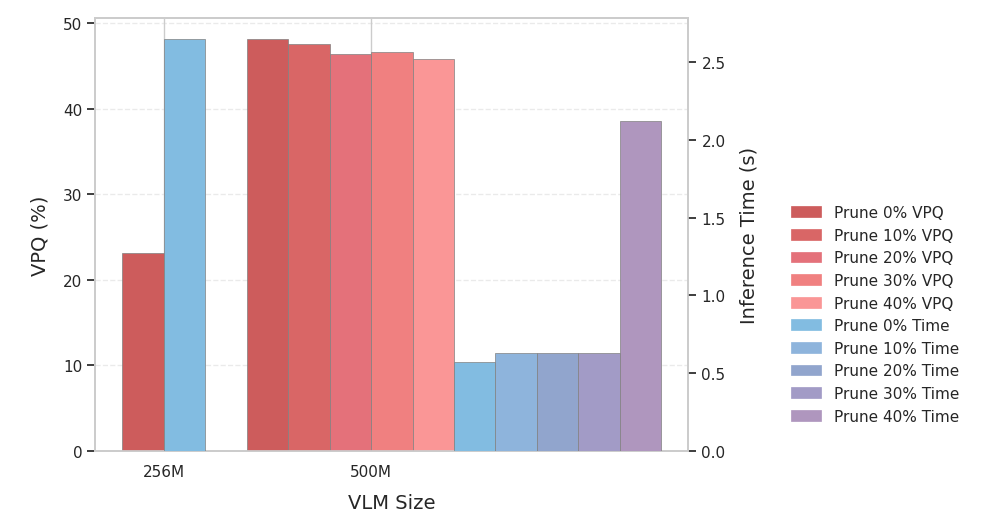}
    \caption{Impact of BEV size and pruning on VPQ and inference time}
    \label{fig:bev_prune_tradeoff}
\end{figure}

\subsubsection{Effect of Model Size and Prune Rate}

Figure~\ref{fig:bev_prune_tradeoff} illustrates the impact of VLM size and pruning rates on REACT's inference time and VPQ performance. The 256M model yields the lowest inference latency but suffers from reduced VPQ, indicating that its limited capacity is insufficient for processing structured prompts and multimodal context effectively. In contrast, the 500M variant maintains consistently high VPQ scores across all pruning levels, demonstrating robust representational capability even under compression.

As the pruning rate increases from 0\% to 40\%, the inference time for the 500M VLM remains relatively stable up to a point, suggesting that moderate pruning has minimal latency impact. However, VPQ performance gradually declines, from 48.5\% at 0\% pruning to 45.5\% at 40\%, highlighting a trade-off between prediction accuracy and computational efficiency. Interestingly, inference latency unexpectedly spikes at the 40\% pruning level, likely due to irregular sparsity patterns and GPU kernel fragmentation, which hinder memory access and parallel execution. This result underscores that overly aggressive pruning can disrupt hardware-level optimizations, leading to diminished performance despite reduced model size.

\begin{table}
    \centering
    \caption{Ablation study on edge adaption strategies. SDPA: Scaled Dot-Product Attention.}
    \label{tab:edge_ablation}
    \begin{tabular}{llllcc}
        \toprule
        \textbf{Quantization} & \textbf{Token Length} & \textbf{BEV Size (Pixel)} & \textbf{Attention Impl.} & \textbf{Inference Time (s)} & \textbf{VPQ (\%)} \\
        \midrule
        \multirow{5}{*}{8-bit}
            & \multirow{2}{*}{\textasciitilde2200} & \multirow{2}{*}{800$\times$800} & SDPA & 1.73 & 49.84\\
            &                                   &                                   & Base & 1.99 & 49.93\\
            & \multirow{3}{*}{\textasciitilde1500} & \multirow{2}{*}{800$\times$800} & SDPA & 0.56& 48.59\\
            &                                   &                                   & Base & 0.57& 48.69\\
            &                                   & 64$\times$64                      & SDPA & 0.55& 48.03\\
            &                                   &                                   & Base & 0.55& 48.04\\
        \midrule
        \multirow{6}{*}{16-bit}
            & \multirow{2}{*}{\textasciitilde2200} & \multirow{2}{*}{800$\times$800} & SDPA & 1.67 & 49.79\\
            &                                   &                                   & Base & 2.03 & 49.71\\
            & \multirow{2}{*}{\textasciitilde1500} & \multirow{2}{*}{800$\times$800} & SDPA & 0.56 & 49.16\\
            &                                   &                                   & Base & 0.60 & 49.16\\
            &                                   & 64$\times$64                      & SDPA & 0.57& 48.24\\
            &                                   &                                   & Base & 0.58& 48.19\\
        \bottomrule
    \end{tabular}
\end{table}

\subsubsection{Effect of Edge Adaption Strategies}

This section investigates how various edge adaptation strategies on AGX affect the runtime efficiency and output quality of the REACT framework. The Table~\ref{tab:edge_ablation} reveals that token length reduction and BEV size compression are the most effective in reducing inference latency with minimal VPQ degradation. While 8-bit quantization generally offers lower latency, certain configurations favor 16-bit due to hardware-specific optimizations. SDPA consistently improves speed but shows diminishing returns under lightweight settings. Based on the trade-off between runtime and predictive quality, the optimal configuration includes 16-bit quantization, a 1500-token input length, a $64 \times 64$ BEV size, and SDPA. The following paragraphs
provide a detailed breakdown of how each individual strategy contributes to this balance.

\textit{Impact of Quantization Precision} \\
8-bit quantization reduces memory footprint and accelerates matrix operations. However, SmolVLM2 is optimized for 16-bit, so under lightweight settings, the 16-bit model can run faster due to more efficient kernel execution.

\textit{Impact of Token Length Reduction} \\
Shorter textual prompts lead to fewer input tokens processed by the transformer layers, reducing self-attention and feedforward computations. This lowers latency significantly from 1.73s to 0.56s, though excessive truncation slightly reduces semantic richness and VPQ.

\textit{Impact of BEV Size Reduction} \\
Smaller BEV maps produce fewer visual tokens, reducing the computational load on the visual encoder and the VLM. Compressing resolution from 800×800 to 64×64 speeds up inference with only minor drops in VPQ, confirming spatial features remain informative post-pooling.

\textit{Impact of Attention Enhancement} \\
SDPA improves attention throughput by optimizing matrix multiplications and memory access. It yields notable gains under high token counts but brings limited benefits when inputs are already small, due to lower GPU utilization.

\begin{table}
    \centering
    \caption{Ablation study on modules. ↑: higher is better; ↓: lower is better. Green: improvement; Red: degradation.}
    \label{tab:mod_ablation}
    \begin{tabular}{>{\raggedright\arraybackslash}p{0.14\linewidth}>{\centering\arraybackslash}p{0.07\linewidth}>{\centering\arraybackslash}p{0.07\linewidth}>{\centering\arraybackslash}p{0.07\linewidth}>{\centering\arraybackslash}p{0.06\linewidth}>{\centering\arraybackslash}p{0.06\linewidth}>{\centering\arraybackslash}p{0.10\linewidth}>{\raggedright\arraybackslash}p{0.10\linewidth}}
        \toprule
         \textbf{Module Removed}&
        \multicolumn{2}{c}{\textbf{minADE ↓ (m)}} & \textbf{FDE ↓ (m)} &
        \multicolumn{2}{c}{\textbf{Motion ↑}} & \textbf{Collision Rate ↓} & \textbf{Avg Inference Time ↓ (s)}\\
          & 1 s & 2 s &  & VPQ & mIOU & & AGX\\
          \midrule
          Task Alignment         
          & 2.125 {\color{red}(+0.604)} 
          & 4.301 {\color{red}(+1.204)} 
          & 3.64 {\color{red}(+1.152)}& 45.9 {\color{red}(-2.3)}& 56.0 {\color{red}(-3.1)} 
          & 26.7 {\color{red}(+14.2)}& 8.95 {\color{red}(+8.38)} \\
          
          RTF                     
          & 3.070 {\color{red}(+1.549)}& 5.820 {\color{red}(+2.723)}& 4.445 {\color{red}(+1.957)}& 43.6 {\color{red}(-4.6)} 
          & 47.5 {\color{red}(-11.6)} 
          & 45.5 {\color{red}(+33.0)}& 3.46 {\color{red}(+2.89)} \\
        \bottomrule
    \end{tabular}
\end{table}

\begin{table}
    \centering
    \caption{Ablation study on inputs. ↑: higher is better; ↓: lower is better. Green: improvement; Red: degradation.}
    \label{tab:input_ablation}
    \begin{tabular}{>{\raggedright\arraybackslash}p{0.14\linewidth}>{\centering\arraybackslash}p{0.07\linewidth}>{\centering\arraybackslash}p{0.07\linewidth}>{\centering\arraybackslash}p{0.07\linewidth}>{\centering\arraybackslash}p{0.06\linewidth}>{\centering\arraybackslash}p{0.06\linewidth}>{\centering\arraybackslash}p{0.10\linewidth}>{\raggedright\arraybackslash}p{0.10\linewidth}}
        \toprule
         \textbf{Inputs Removed}&
        \multicolumn{2}{c}{\textbf{minADE ↓ (m)}} & \textbf{FDE ↓ (m)} &
        \multicolumn{2}{c}{\textbf{Motion ↑}} & \textbf{Collision Rate ↓} & \textbf{Avg Inference Time ↓ (s)}\\
          & 1 s & 2 s &  & VPQ & mIOU & & AGX\\
        \midrule
         Camera Frame            
         & 1.901 {\color{red}(+0.380)}& 3.904 {\color{red}(+0.807)}& 3.20 {\color{red}(+0.712)}& 45.0 {\color{red}(-3.2)} 
         & 55.8 {\color{red}(-3.3)} 
         & 36.5 {\color{red}(+24.0)}& 0.51 {\color{green}(-0.06)}\\
         
         RSU Hazard              
         & 2.303 {\color{red}(+0.782)}& 4.882 {\color{red}(+1.785)}& 3.90 {\color{red}(+1.412)}& 44.0 {\color{red}(-4.2)} 
         & 53.0 {\color{red}(-6.1)} 
         & 30.0 {\color{red}(+17.5)}& 0.56 {\color{green}(-0.01)}\\
         
         Navigation Information  
         & 1.755 {\color{red}(+0.234)}& 3.490 {\color{red}(+0.393)}& 2.93 {\color{red}(+0.442)}& 46.2 {\color{red}(-2.0)} 
         & 57.2 {\color{red}(-1.9)} 
         & 17.0 {\color{red}(+4.5)}& 0.57 \\

         Ego Status              
         & 1.615 {\color{red}(+0.094)}& 3.363 {\color{red}(+0.266)}& 2.71 {\color{red}(+0.222)}& 45.4 {\color{red}(-2.8)} 
         & 54.5 {\color{red}(-4.6)} 
         & 15.0 {\color{red}(+2.5)}& 0.56 {\color{green}(-0.01)}\\
        \bottomrule
    \end{tabular}
\end{table}

\subsubsection{Effect of Inputs and Modules}
\textit{Visual Input} \\
The ablation results in Table~\ref{tab:input_ablation} underscore the importance of each input stream and processing module in REACT. Removing the camera frame input significantly degrades performance across all metrics. Without visual inputs, REACT loses critical contextual information, making it unable to localize predictions accurately, such as aligning trajectories with lanes or positioning vehicles behind others appropriately. This results in substantial increases in minADE and FDE, and the collision rate rises sharply to 36.5\%, confirming that visual context from the BEV feature map is vital for spatial reasoning and motion safety. Although inference time is reduced when BEV features are excluded from the VLM input, the accompanying performance degradation shows that this is not a viable direction for optimization. The trade-off strongly favors retaining visual input to preserve safety and prediction accuracy.

\textit{RSU Hazard Input} \\
Excluding the RSU hazard input also results in a noticeable performance decline, particularly in collision rate and long-horizon displacement errors. This highlights the importance of infrastructure, providing hazard information for anticipating downstream risks. Without RSU alerts, the vehicle reverts to relying solely on onboard perception and is unable to proactively adjust its trajectory to avoid imminent hazards in advance.

\textit{Navigation Information Input} \\
Ablating the navigation information leads to moderate increases in displacement errors and a slight decline in motion prediction metrics. However, the collision rate remains relatively low at 17.0\%, suggesting that REACT can still operate safely with limited routing context, though with reduced precision. Notably, without navigation data, REACT fails to produce appropriate stopping behavior behind stop bars at intersections, indicating its reliance on this input for fine-grained trajectory alignment with road semantics.

\textit{Ego Status Input} \\
When the ego status input is removed, REACT no longer receives real-time information about the vehicle’s current speed and position. This omission also affects the interpretation of visual inputs, although REACT is still able to infer future positions to a limited extent using contextual cues. Both minADE and FDE show slight increases, and overall motion quality declines modestly. Nevertheless, the safety impact is relatively minor, implying that ego-centric kinematic inputs serve a supportive, but not dominant, role in REACT’s trajectory reasoning.

\textit{Task Projection} \\
Table~\ref{tab:mod_ablation} shows that disabling the task projection module causes a substantial increase in inference time, as the structured task representation is no longer available to guide the LLM efficiently. This results in higher displacement errors and a collision rate of 26.7\%, underscoring the importance of semantically organizing input data before reasoning. The results affirm that structured prompts not only improve reasoning efficiency but also contribute directly to prediction quality and safety.

\textit{RTF} \\
Table~\ref{tab:mod_ablation} demonstrates the benefits of the RTF module. Without RTF, the model must predict absolute future waypoint positions, increasing the output token length and inference time. This direct prediction approach also degrades accuracy, significantly raising both average and final displacement errors. Additionally, collision rates increase in the absence of RTF, indicating reduced robustness in safety-critical scenarios. By shifting the prediction task to residual adjustments, RTF not only reduces decoding complexity but also leverages prior trajectory context to enhance both runtime efficiency and safety performance.

\section{Conclusion and Discussion}

This paper introduces REACT, a real-time, V2X-integrated edge VLM framework specifically designed for invisible hazard avoidance. By combining a lightweight VLM, structured prompt engineering, CoT SFT, and RTF, REACT demonstrates strong performance across key dimensions including contextual reasoning, multimodal data fusion, real-time edge computing, generalization, and safety-focused trajectory planning.

Experimental evaluations confirm REACT’s effectiveness, achieving a mIOU of 59.1\%, and VPQ of 48.2\%. Compared to existing non-GPT Transformer-based planners, such as CoBEVT and V2X-ViT, REACT provides significant improvements in both safety and prediction accuracy. Moreover, it attains an average CRR of 77.0\% and maintains an adaptive MCD ranging between 1.09 and 6.24 meters, highlighting its robust risk awareness. REACT also demonstrates resilience to varying environmental conditions, maintaining consistent performance across different weather and lighting settings. Its efficient edge implementation achieves an inference latency of 0.57 seconds on AGX, confirming practical viability for real-time deployment on resource-constrained devices. Key findings across specific capabilities are summarized below:

\textit{Contextual Reasoning} \\
REACT surpasses baselines by leveraging multi-source information and a finetuned VLM, effectively handling occlusions and multi-agent interactions. Both qualitative examples and quantitative results confirm its proactive risk anticipation and appropriate trajectory adjustments.

\textit{Multimodal Fusion} \\
By integrating BEV visual features and symbolic inputs, REACT achieves superior VPQ and mIOU scores. Ablation studies highlight the necessity of both visual and symbolic modalities for comprehensive spatial and contextual reasoning.

\textit{Edge Deployment} \\
Optimizations including offline quantization, attention enhancement, and token length reduction allow REACT to operate effectively on Jetson-class edge hardware. The strategies specifically provide REACT with an optimal balance between prediction accuracy and real-time performance, enabling practical edge deployment.

\textit{Generalization} \\
Through multimodal inputs and structured reasoning, REACT exhibits robust generalization capabilities. It maintains reliable performance in various or unseen scenarios, with minimal degradation compared to baselines, underscoring its adaptability and resilience.

Nevertheless, this research identifies several limitations for potential refinement. Current VLM predictions show less precision in yaw angle estimation compared to positional accuracy, occasionally resulting in misaligned orientations. Additionally, due to prompt brevity necessitated by computational constraints, predicted positions may sometimes inaccurately overlap sidewalks or adjacent agents. Furthermore, REACT’s conservative nature occasionally results in overly cautious stopping distances, particularly in ambiguous or low-risk situations.

Future research should aim to improve the precision of yaw angle and speed estimation, enhance trajectory predictions in complex spatial environments, and broaden the framework’s applicability to other types of abnormal behaviors exhibited by VRUs and surrounding vehicles. Further advancements may include deeper model compression, extended prediction horizons, and integrating richer contextual inputs such as detailed road topology and explicit vehicle intent.  Advancing these aspects will further solidify REACT’s role as a robust foundation for next-generation autonomous driving systems, ones that prioritize safety, efficiency, adaptability, and intelligent context understanding at the edge.

\section{Funding}

The authors disclosed no financial support for the research, authorship, and/or publication of this article.

\section{Declaration of Generative AI and AI-assisted Technologies in the Writing Process}

During the preparation of this work the author(s) used OpenAI’s GPT-4o language model solely in order to polish grammar and improve the clarity of writing during manuscript preparation. After using this tool/service, the author(s) reviewed and edited the content as needed and take(s) full responsibility for the content of the published article.

\section{Author Contributions}

Fengze Yang led the conceptualization, methodology design, software development, and original draft writing. Bo Yu conducted the experiments and prepared the baseline models based on the dataset. Zhou Yang contributed to idea refinement, paper structuring, and writing. Zhengzhong Tu contributed to model fine-tuning, data processing, and manuscript writing. Xuewen Luo contributed to the literature review, data processing, and manuscript polishing. Chenxi (Dylan) Liu (corresponding author) oversaw project administration and supervision, and contributed to manuscript review and editing.

\section{Declaration of Conflicting Interests}
All authors declared no potential conflicts of interest with respect to the research, authorship, and/or publication of this article.

\bibliographystyle{ieeetr}
\bibliography{1-references}

\end{document}